%% file: acl_latex.tex
\documentclass[11pt]{article}
\usepackage{xspace}
\usepackage[final]{acl}

\usepackage{times}
\usepackage{latexsym}
\usepackage{amsthm}
\usepackage{amsmath}
\usepackage{amsfonts}
\usepackage{amssymb}
\usepackage{bbm}
\usepackage{multirow}
\usepackage{multicol}
\usepackage{algorithm}
\usepackage{algpseudocode}

\usepackage{booktabs}
\usepackage{pifont}
\newcommand{\xmark}{\ding{55}} 

\usepackage[T1]{fontenc}

\usepackage[utf8]{inputenc}

\usepackage{microtype}

\usepackage{inconsolata}

\usepackage{graphicx}

\usepackage[most]{tcolorbox}

\definecolor{qback}{HTML}{F7FAFC}   
\definecolor{qframe}{HTML}{CBD5E1}  
\definecolor{aback}{HTML}{EFF6FF}   
\definecolor{aframe}{HTML}{93C5FD}  
\definecolor{jback}{HTML}{FFF7ED}   
\definecolor{jframe}{HTML}{FDBA74}  
\definecolor{okbg}{HTML}{DCFCE7}    
\definecolor{okfg}{HTML}{166534}
\definecolor{badbg}{HTML}{FEE2E2}   
\definecolor{badfg}{HTML}{7F1D1D}

\tcbset{
  breakable,
  sharp corners,
  boxrule=0.6pt,
  left=8pt,right=8pt,top=8pt,bottom=8pt,
  coltitle=black,
  fonttitle=\bfseries,
  before skip=8pt, after skip=10pt,
}

\newtcolorbox[use counter=promptbox]{PromptBox}[1][]{%
  colback=aback,
  colframe=aframe,
  title=Prompt,
  #1
}

\definecolor{trailRL}{HTML}{FF9900}       
\definecolor{trailSFT}{HTML}{CC7A00}      
\definecolor{trailNoReward}{HTML}{FFB84D} 
\definecolor{qwenAgent}{HTML}{0072B2}     
\definecolor{sonnetAgent}{HTML}{56B4E9}   

\title{SQL-Trail: Multi-Turn Reinforcement Learning with Interleaved Feedback for Text-to-SQL} 

\author{
 \textbf{Harper Hua\textsuperscript{1}\thanks{Work done during internship at AWS.}},
 \textbf{Zhen Han\textsuperscript{2}},
 \textbf{Zhengyuan Shen\textsuperscript{2}},
 \textbf{Jeremy Lee\textsuperscript{2}},
\\
 \textbf{Patrick Guan\textsuperscript{2}},
 \textbf{Qi Zhu\textsuperscript{2}},
 \textbf{Sullam Jeoung\textsuperscript{2}},
 \textbf{Yueyan Chen \textsuperscript{2}},
\\
 \textbf{Yunfei Bai\textsuperscript{2}},
 \textbf{Shuai Wang\textsuperscript{2}},
 \textbf{Vassilis Ioannidis\textsuperscript{2}},
 \textbf{Huzefa Rangwala\textsuperscript{2}}
\\
 \textsuperscript{1}Stanford University,
 \textsuperscript{2}Amazon Web Services
}

\definecolor{sqlblue}{RGB}{0,70,200}
\definecolor{obsorange}{RGB}{200,90,0}
\definecolor{solgreen}{RGB}{0,150,60}
\definecolor{thinkpurple}{RGB}{140,0,180}

\newcommand{\method}{\textsc{\textbf{SQL-Trail}}\xspace}

\newcommand{\sql}{\texttt{\textcolor{sqlblue}{<sql>}}}
\newcommand{\sqlend}{\texttt{\textcolor{sqlblue}{</sql>}}}

\newcommand{\obs}{\texttt{\textcolor{obsorange}{<observation>}}}
\newcommand{\obsend}{\texttt{\textcolor{obsorange}{</observation>}}}

\newcommand{\sol}{\texttt{\textcolor{solgreen}{<solution>}}}
\newcommand{\solend}{\texttt{\textcolor{solgreen}{</solution>}}}

\newcommand{\think}{\texttt{\textcolor{thinkpurple}{<reasoning>}}}
\newcommand{\thinkend}{\texttt{\textcolor{thinkpurple}{</reasoning>}}}

\begin{document}
\maketitle
\begin{abstract}
  While large language models (LLMs) have substantially improved Text-to-SQL generation, a pronounced gap remains between AI systems and human experts on challenging benchmarks such as BIRD-SQL. We argue this gap stems largely from the prevailing single-pass paradigm, which lacks the iterative reasoning, schema exploration, and error-correction behaviors that humans naturally employ. To address this limitation, we introduce \method, a multi-turn reinforcement learning (RL) agentic framework for Text-to-SQL. Rather than producing a query in one shot, \method interacts with the database environment and uses execution feedback to iteratively refine its predictions. Our approach centers on two key ideas: (i) an adaptive turn-budget allocation mechanism that scales the agent’s interaction depth to match question difficulty, and (ii) a composite reward panel that jointly incentivizes SQL correctness and efficient exploration. Across benchmarks, \method sets a new state of the art and delivers strong data efficiency—up to \textbf{18×} higher than prior single-pass RL state-of-the-art methods. Notably, our 7B and 14B models outperform substantially larger proprietary systems by \textbf{5\%} on average, underscoring the effectiveness of interactive, agentic workflows for robust Text-to-SQL generation.
\end{abstract}

\input{sections/introduction}

\input{sections/related_work}
\input{sections/method}
\input{sections/experiments}

\input{sections/results}
\input{sections/conclusion}


\section{Limitations}
Despite strong gains, \method has several important limitations. First, it assumes an interactive execution environment: the agent must be able to run (possibly multiple) SQL queries against the target database to obtain errors and results. This requirement may be infeasible in settings with restricted connectivity, strict privacy controls, or expensive query execution. Second, multi-turn interaction increases inference cost and latency (more tokens, more database calls). While difficulty-aware turn budgeting reduces unnecessary steps, worst-case overhead remains higher than single-pass systems. Third, parts of the training recipe rely on supervision signals that may not be available at scale in new domains. Reward shaping can also introduce inductive biases (e.g., favoring syntactic/structural similarity over alternative but equivalent SQL), and RL optimization may exploit spurious correlations in the training distribution. Finally, our empirical study is centered on Spider/BIRD and related robustness suites; these benchmarks do not fully represent production constraints and real-world use cases. As a result, additional validation is needed to establish reliability and cost-quality trade-offs in deployed environments.

\bibliography{custom}

\clearpage
\appendix
\section{Appendix}
\label{sec:appendix}
\input{sections/Appendix}

\end{document}

%% file: sections/introduction.tex
\section{Introduction}
\begin{figure}
    \centering
    \includegraphics[width=1\linewidth]{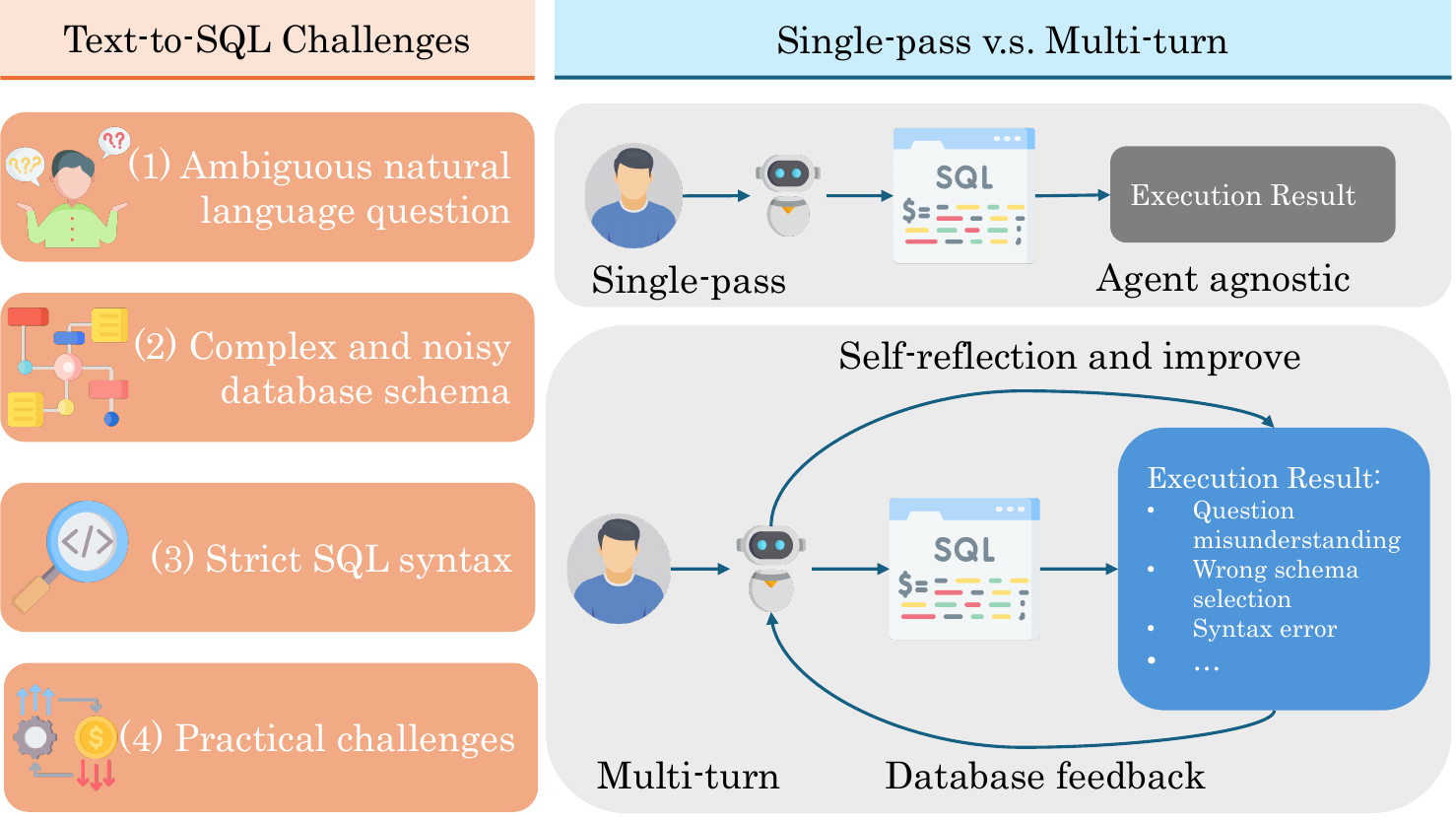}
    \caption{Illustration of core Text-to-SQL challenges and the shift from single-pass to multi-turn generation.}
    \label{fig:challenges}
\end{figure}

Text-to-SQL enables intuitive access to structured databases by automatically converting natural language questions into executable SQL queries, thereby democratizing data retrieval for non-expert users~\citep{qin2022survey, liu2025survey}. Recent large language models (LLMs) have achieved remarkable progress in this task, advancing both research frontiers and real-world applications~\citep{openaigpt4, anthropic_claude_3.7_sonnet_system_card}. Current state-of-the-art approaches for LLM adaptation primarily rely on chain-of-thought prompting~\citep{dinsql, c3sql, petsql}, supervised fine-tuning (SFT)~\citep{omnisql, macsql}, or reinforcement learning (RL)~\citep{ma2025sqlr1trainingnaturallanguage, reasoningsql}. However, despite these sophisticated methods, a substantial performance gap remains between humans and the best AI systems on challenging leaderboards such as BIRD-SQL~\cite{blier2021unbiased}.

This gap largely stems from the single-pass paradigm underlying most existing methods: given a natural language question and a database schema, the model directly generates a SQL query without leveraging feedback from the database environment, such as execution results or error messages, into its reasoning process~\citep{chen2025text, elgohary-etal-2020-speak}. Moreover, as shown in Figure~\ref{fig:challenges}, Text-to-SQL poses several intrinsic challenges: (1) natural language questions are often ambiguous~; (2) database schemas can be large, complex, and filled with noisy or semantically ambiguous entity names; and (3) SQL’s strict syntactic structure leaves little room for error; (4) practical constraints regarding high computational costs and the scarcity of large-scale, high-quality public training datasets~\citep{liu2025survey}. As noted by \citet{pourreza2024din}, most model failures arise from incorrect schema linking, namely difficulty identifying the most relevant tables and columns, and from hard examples involving nested subqueries, multi-hop joins, and complex aggregations. In contrast, human experts excel at this task because they can iteratively interact with the database to better understand its structure, decompose complex problems into smaller subqueries to test intermediate results, and refine or debug their queries based on execution feedback~\citep{liu2025survey}. 

Recent work on tool-augmented RL has shown that LLM agents can improve substantially by interacting with external environments over multiple turns (e.g., search, UI, code execution) rather than acting in a single pass~\citep{searchr1, wei2025swe, hu-etal-2025-os, wei2025webagentr1trainingwebagents}. Such interaction enables agents to gather missing information, refine intermediate hypotheses, and self-correct using feedback, yielding better accuracy and robustness across domains~\citep{liu2025skyrlsql, li2025flow, cao2025skyrl}. Motivated by this paradigm, we propose \method, a multi-turn RL-trained Text-to-SQL agent. Unlike single-pass RL baselines such as SQL-R1~\citep{ma2025sqlr1trainingnaturallanguage}, \method performs iterative database probing, schema exploration, and execution-based self-correction within a closed loop, with difficulty-aware turn budgeting to avoid overthinking on easy queries while allocating more interaction steps to harder ones. We further introduce a composite reward panel that provides dense step-wise guidance, jointly encouraging execution correctness and efficient long-horizon behavior, enabling smaller models to reliably converge to accurate SQL programs.

Our main contributions are as follows: 
(1) \textbf{Unified Multi-turn RL Training Framework:} We present the first end-to-end study of multi-turn LLM training strategies for Text-to-SQL, integrating a novel adaptive turn-budget allocation mechanism that allows the agent to expend more interaction turns on complex queries while remaining concise on simple ones. 
(2) \textbf{Systematic Investigation of Multi-turn Agent Behavior:} We provide a comprehensive comparison between single-pass and multi-turn RL and detailed ablation study of our composite reward panel, revealing how interactive execution fundamentally reshapes reasoning trajectories. We also conduct the first systematic investigation into agent reasoning efficiency under a multi-turn setting.
(3) \textbf{State-of-the-Art Data Efficiency and Generalization:} Our experiments demonstrate that multi-turn RL achieves state-of-the-art out-of-distribution performance with exceptional data efficiency. Specifically, while prior single-pass methods such as OmniSQL yield a marginal accuracy improvement of 0.005\% per 1,000 in-distribution training samples, \method delivers a 4.6\% gain per 1,000 samples using out-of-distribution data. This high-potential learning paradigm enables our 7B model to outperform significantly larger proprietary models by an average of 5\% across all benchmarks, establishing new performance records for open-source models at both the 7B and 14B scales.

%% file: sections/related_work.tex
\section{Related Work}
Text-to-SQL has progressed from rule-based and template systems to neural semantic parsers such as Seq2SQL~\citep{zhong2017seq2sql} and SQLNet~\citep{xu2017sqlnet}, and more recently to LLM-based approaches. Modern methods leverage in-context learning~\citep{zhang2023actsqlincontextlearningtexttosql, agarwal2024manyshot, sun2023sqlprompt}, improved schema linking~\citep{li2023resdsql, cao2024rsl, snell2024scaling, eyal2023semantic}, and constrained decoding (e.g., PICARD)~\citep{scholak2021picard} to improve validity and robustness. State-of-the-art systems largely rely on supervised fine-tuning and multi-step self-correction with execution feedback~\citep{omnisql, wang2023mac, chasesql, pourreza2024dts}, but SFT-centric training can still struggle to generalize to unseen schemas and complex queries~\citep{reasoningsql, ma2025sqlr1trainingnaturallanguage}.

In parallel, reinforcement learning has emerged as a powerful mechanism for training autonomous agents capable of complex, multi-turn reasoning~\citep{wei2022chain, jaech2024openai, plaat2024reasoning}. Recent frameworks such as DeepSeek-R1~\citep{deepseekr1} have demonstrated that algorithms like Group Relative Policy Optimization (GRPO) can incentivize models to internalize intermediate reasoning steps without explicit supervision, achieving strong performance in mathematics and logic. Furthermore, domain-specific implementations like Llama3-SWE-RL~\citep{wei2025swe} and Search-R1~\citep{searchr1} have proven that RL agents can self-improve on real-world software engineering tasks and tool-augmented retrieval by interacting dynamically with their environments~\citep{li2025flow, deepretrieval, zheng2025deepresearcherscalingdeepresearch}. Despite these advancements in general code generation and agentic workflows, there remains a notable gap in applying deep reasoning RL frameworks to the multi-turn Text-to-SQL setting~\citep{liu2025survey}.



%% file: sections/method.tex
\section{\method}
\begin{figure*}[t]
  \includegraphics[width=1.0\linewidth]{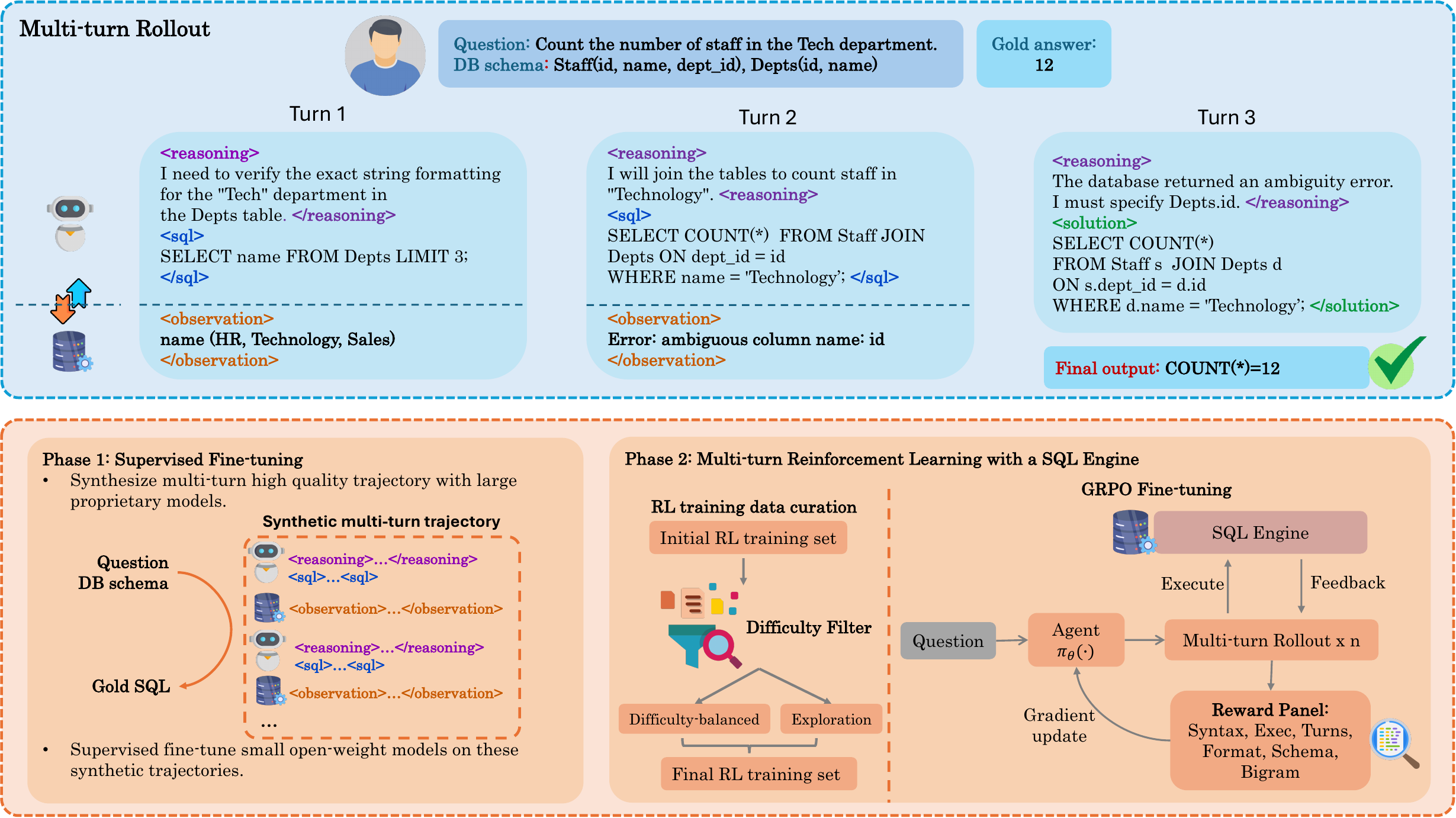}
  \caption{Overview of \method. The top panel demonstrates the multi-turn Text-to-SQL interaction process, and the bottom panel outlines the unified RL training pipeline.}
\end{figure*}
\paragraph{Overview}

We introduce \method, a multi-turn Text-to-SQL agent fine-tuned through a staged training pipeline. 

\subsection{Generation with Multi-Turn SQL Engine Calling}\label{sec:3.1}
We design a tool-augmented ReAct-style agent workflow~\citep{yao2023react} in which an LLM agent interacts with the database environment by calling a SQL execution tool and receives structured feedback from the database, enabling it to solve Text-to-SQL tasks through multi-turn reasoning and execution. The framework consists of two core components: \emph{actions}, where the LLM produces a reasoning trace and proposes SQL queries, and \emph{observations}, which are the execution outputs returned to the model.

Formally, for the $i$-th example, we represent the multi-turn interaction as a trajectory
\begin{equation}
\tau_i \;=\; \bigl\{(o_{i,t}, a_{i,t})\bigr\}_{t=1}^{T_i},
\end{equation}
where $a_{i,t}$ denotes the agent action at turn $t$ (e.g., a reasoning trace and a SQL query to execute), and $o_{i,t}$ denotes the corresponding observation returned by the environment (e.g., execution results, error messages, or a truncated dataframe preview). Here $T_i$ is the number of interaction turns for instance $i$. A trajectory is thus the ordered sequence of alternating actions and observations induced by iterative SQL engine calls.

Our approach adopts an iterative reason–execute–observe loop, where the LLM alternates between natural-language reasoning and external SQL execution under a strict, token-delimited interface~\citep{cao2025skyrl}. At each turn, the model outputs a reasoning block \think{}...\thinkend{} followed by a SQL action \sql{}...\sqlend{}; the system extracts and executes the SQL, then appends the engine output as an observation \obs{}...\obsend{} for the next turn. The process repeats until a turn budget is reached or the model emits the final solution in \sol{}...\solend{}. Full system prompt details and formatting specifications are provided in Appendix~\ref{app:agent_workflow}.


\subsection{Training phase 1: Supervised Fine-Tuning}
Supervised fine-tuning (SFT) provides a crucial initialization step for our agent. Smaller open-weight models often struggle with the long, highly structured prompts required by the above multi-turn agent workflows, frequently producing formatting mistakes, prematurely invoking tools, or entering undesired action loops~\citep{qwen2.5}. In contrast, large proprietary models demonstrate strong instruction-following and reliably adhere to the agent interface~\citep{anthropic_claude_3.7_sonnet_system_card}. To bridge this gap, we distill the instruction-following behavior of a high-capacity closed-source teacher model into smaller open-source student models, teaching them to execute the agent loop correctly and consistently. This distillation equips the students with a strong prior over the agent’s operational structure, enabling stable early rollouts and providing dense formative signals that significantly improve downstream reinforcement learning.

\subsection{Training Phase 2: Multi-turn Reinforcement Learning with a SQL Engine}

In the second training phase, we apply reinforcement learning (RL) to optimize the LLM as a flexible multi-turn agent that develops a deeper understanding of how to use the SQL execution tool to solve tasks precisely and efficiently. RL encourages the agent to avoid unnecessary detours and to generalize more robustly under distribution shifts in database schemas and unseen domains. To train this multi-turn Text-to-SQL agent, we adopt an RL framework built upon Grouped Reinforcement Policy Optimization (GRPO)~\citep{shao2024deepseekmath} and extend it with a detailed reward panel tailored for multi-turn reasoning. 

\subsubsection{Reinforcement Learning Formulation}
For each training instance sampled from the dataset $\mathcal{D}$, the input consists of a natural language question $q$ and its associated database schema $d$. 




We distinguish \emph{single-turn} RL, where the policy emits one SQL query in a single step and receives a terminal execution-based reward, from our \emph{multi-turn} RL setting, where the policy interacts with a SQL execution environment over multiple steps. In the multi-turn case, conditioned on $(q,d)$, the rollout engine (Section~\ref{sec:3.1}) induces a trajectory $\tau={(o_t,a_t)}_{t=1}^{T}$, where each action $a_t$ proposes a SQL (or control) step and each observation $o_t$ is structured engine feedback (e.g., results or errors); learning is driven by a trajectory-level reward $R(\tau;q,d)$ that combines final execution correctness with intermediate behavioral signals.

To better encourage exploration in the multi-turn setting, we remove the KL regularization term and modify the clipping mechanism used in the original GRPO. Standard GRPO applies symmetric PPO-style clipping~\citep{yu2025dapo}, enforcing identical upper and lower bounds on the likelihood ratio $
\rho_i(\theta) =
\frac{\pi_\theta(\tau_i \mid q,d)}{\pi_{\theta_{\text{old}}}(\tau_i \mid q,d)}
$, where $\pi_{\theta_{\text{old}}}$ is the policy before the update step. In contrast, we adopt a clip-higher variant, which keeps the conservative lower bound to maintain training stability but raises the upper bound to allow larger update steps on promising but initially low-likelihood trajectories. Formally, we constrain $\rho_i(\theta)$ in between $1-\epsilon_{\text{low}}$ and  $1-\epsilon_{\text{high}}$ with $\epsilon_{\text{low}}<\epsilon_{\text{high}}$. By expanding the upper clipping range, the clip-higher strategy increases probability mass on diverse exploratory rollouts while still maintaining controlled updates.

Our objective maximizes the modified GRPO return:
\begin{equation}\small
\begin{aligned}
&\mathcal{J}_{GRPO}(\theta) \\
&=\mathbb{E}_{q,d\sim\mathcal{D}, \{\tau_i\}^G_{i=1}\sim \pi_{\theta_{old}}(\cdot|q,d)}\!\left[
\frac{1}{G}\sum_{i=1}^{G}
\min\!\Bigg(
\rho_i(\theta) A_i,\;
\right. \\
&\qquad\qquad\left.
\mathrm{clip}\!\big(
\rho_i(\theta),\;
1-\epsilon_{\text{low}},\;
1+\epsilon_{\text{high}}
\big) A_i
\Bigg)
\right].
\end{aligned}
\end{equation}

This formulation preserves GRPO’s group-relative optimization
while providing stronger encouragement for upward, exploration-driven policy updates.


\subsubsection{Multi-turn Reward Design}\label{sec:reward}
Reward design is the main driver of effective RL. Prior multi-turn agent RL training largely rely on a binary execution reward~\citep{liu2025skyrlsql, searchr1}, but we find this signal is overly sparse: small mistakes collapse the reward to zero and offer little direction for improvement, especially when training data is limited. To better exploit the available data, we introduce a six-term, rule-based reward panel that provides fine-grained SQL-structural feedback and shapes long-horizon multi-turn behavior.






\paragraph{Final execution reward.}
Our primary objective is execution correctness, so we use a binary execution reward $r_{\text{exec}}$: we extract the final SQL from \sol{}...\solend{}, execute it, and compare the result to the gold query’s execution result,
\begin{equation}
    r_{\text{exec}} = \mathbbm{1}[\text{exec}(\text{pred\_sql}) = \text{exec}(\text{gold\_sql})]
\end{equation}

\paragraph{Turn number reward.}
To encourage efficient multi-turn behavior, we add a turn-budget reward $r_{\text{turns}}$ that favors solving within a small number of steps (with thresholds tied to difficulty) and discourages redundant interactions. The exact thresholding rule is provided in Appendix~\ref{app:reward_panel}.

\paragraph{Auxiliary shaping rewards.}
We further include four lightweight rewards to densify supervision and enforce valid behavior: bi-gram overlap ($r_{\text{bigram}}$) and schema grounding ($r_{\text{schema}}$) as Jaccard-based similarity signals, plus binary syntax validity ($r_{\text{syntax}}$) and interface compliance ($r_{\text{format}}$). Full definitions and implementation details are deferred to Appendix~\ref{app:reward_panel}.

\paragraph{Total reward.}
We combine terms with a simple weighted sum,
\begin{equation}\small
R = 5r_{\text{exec}} + 2r_{\text{turns}} + r_{\text{schema}} + r_{\text{bigram}} + r_{\text{syntax}} + r_{\text{format}},
\end{equation}
using larger weights on $r_{\text{exec}}$ (end-task objective) and $r_{\text{turns}}$ (explicitly shaping multi-turn efficiency), while keeping auxiliary shaping terms at unit weight (stabilization without dominating optimization). The weighting rationale and ablations are included in Appendix~\ref{app:reward_panel}.

%% file: sections/experiments.tex
\section{Experiment Setup}

\subsection{Model setup}
We use Qwen2.5-Coder base models (3B/7B/14B)~\citep{hui2024qwen2} following existing works and a two-stage pipeline: supervised fine-tuning (SFT) followed by RL from the SFT checkpoints. For RL, we sample $G=6$ rollouts per query (temperature 1.0) with a 10-turn trajectory cap to enable iterative planning and revision. Additional training details are provided in Appendix~\ref{app:train_confg}.

\subsection{Training data curation}
We use a two-stage pipeline (SFT $\rightarrow$ RL) and deliberately curate a small but informative training set to study data efficiency. Full sampling, filtering, and hyperparameter details are provided in Appendix~\ref{app:data_curation}.
\paragraph{SFT.}
We sample 3,000 Spider-train questions~\citep{yu2018spider}, generate multi-turn trajectories with Claude-Sonnet-3.7 using our agent template (Section~\ref{sec:3.1}), and retain 1,000 trajectories with correct final SQL, priortized toward medium/hard difficulty. These demonstrations distill instruction-following and multi-turn behaviors into the student models. Appendix~\ref{app:data_curation} lists the exact selection criteria and prompts.
\paragraph{RL.}
We train on 1,027 questions split into: (i) a \textbf{difficulty-balanced set} (700) chosen to maximize informative GRPO advantages by favoring “hard-but-solvable” items (non-degenerate pass@6), and (ii) an \textbf{exploration set} (327) of consistently difficult cases to target persistent failure modes (including post-SFT failures and pass@6=0 subsets from SynSQL and Spider). Appendix~\ref{app:data_curation} details the candidate pools, pass@6 estimation, scoring/ranking, and composition breakdown.

\subsection{Benchmarks and Evaluation}

\paragraph{Benchmarks and Metrics.}
We evaluate on Spider~\citep{yu2018spider} and BIRD~\citep{li2024can}. For Spider, we report dev and test results to measure in-domain gains, since they match our SFT training distribution; for out-of-distribution generalization, we evaluate on the BIRD dev set, which is held out from all training phases and contains complex, value-centric questions over real-world schemas across 37 domains. We further assess robustness on Spider-DK~\citep{gan2021exploring}, Spider-Syn~\citep{gan-etal-2021-towards}, and Spider-Realistic~\citep{deng-etal-2021-structure}, which probe external-knowledge reliance, lexical shift, and more ambiguous user queries. We report Execution Accuracy (EX), i.e., the fraction of predictions whose execution result matches the ground truth, and measure training efficiency via Data Efficiency, defined as EX gain (percentage points) per 1,000 training examples:
\begin{equation}
 \text{efficiency}_{\text{pp}/1\text{k}} = \frac{\Delta \text{pp}}{N} \times 1000
\end{equation}
where $\Delta \text{EX}_{\text{pp}}$ is the EX improvement in pp and $N$ is the number of training examples.

\paragraph{Inference Strategy.}
To ensure a fair comparison with prior work, we follow standard test-time evaluation practices: greedy decoding (Pass@1, zero-temperature) and majority voting over 8 sampled candidates using execution-result consensus. We also report Pass@$k$ to characterize the model’s best-of-$k$ capability.

%% file: sections/results.tex
\section{Results}
\input{tables/main_table}
\subsection{Main results}
\paragraph{Baseline Comparison}
Table~\ref{tab:main_results} reports the performance of \method across Spider and BIRD benchmarks, compared against several state-of-the-art Text-to-SQL finetuning systems. We include strong single-pass RL baselines such as SQL-R1~\citep{ma2025sqlr1trainingnaturallanguage}, trained on 5k synthetic challenging SynSQL~\citep{omnisql} examples, and Reasoning-SQL~\citep{reasoningsql}, trained on over 8k BIRD-train samples. We further compare to OminiSQL~\citep{omnisql}, which performs large-scale SFT on 2.5M synthetic examples and chain-of-thought augmented Spider and BIRD datasets.

Overall, \method reaches state-of-the-art accuracy with an order-of-magnitude gain in data efficiency, trained on just 1,873 examples. On BIRD-dev, \method-7B achieves 60.1\% (Greedy) and 64.2\% (Majority), outperforming the much larger Sonnet-3.7 by +1.6 and +4.1 points under identical evaluation. Compared with same-scale single-pass RL baselines (SQL-R1, Reasoning-SQL), \method is far more data-efficient: at 7B, it delivers 7–18× higher efficiency on both Spider-test and BIRD-dev.

\paragraph{Generalization and robustness analysis}
Multi-turn RL substantially improves out-of-distribution generalization. Reasoning-SQL-7B trains on \(>8\)k in-domain BIRD examples yet transfers poorly to Spider-test (78.7\% EX), \(9.2\) points below OminiSQL-7B (87.9\%). In contrast, \method-7B is trained only on Spider + SynSQL (never on BIRD) but transfers strongly to BIRD-dev, trailing OminiSQL-7B by just 3.9 points (64.2\% vs.\ 68.1\%). This asymmetry suggests BIRD supervision does not reliably transfer to Spider, whereas our multi-turn agent trained on Spider transfers robustly to BIRD.

Against SQL-R1-7B (same SynSQL source), \method-7B is more cross-domain robust on BIRD-dev (+3.7 Greedy, +1.1 Majority) while using less than half the data. Compared with OminiSQL---which trains on both Spider and BIRD---\method still leads by \(>5\) points on Spider-dev at 7B (86.8\% vs.\ 81.6\%) and 14B (87.1\% vs.\ 82.0\%), and matches or exceeds OminiSQL on BIRD-dev at the 14B scale despite BIRD being unseen for \method. We attribute this transferability to active environment probing, which helps the agent infer schema structure and user intent across unseen domains.

For robustness, \method-7B matches or outperforms other 7B baselines on Spider-DK, Spider-Syn, and Spider-Realistic (Appendix~\ref{app:single_multi}, Table~\ref{table:robust}), with the largest gains on Spider-Syn, reinforcing the advantage of multi-turn RL under schema perturbations and noisy queries.
\paragraph{Single-pass and Multi-turn RL comparison}
\input{tables/single_vs_multi}
\begin{figure}[t]
    \centering
    
    \includegraphics[width=\columnwidth]{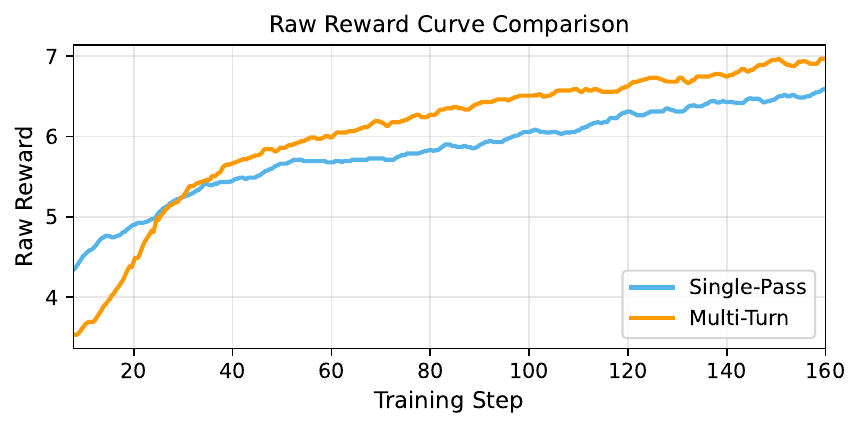}
    \vspace{-25pt}
    
    \includegraphics[width=\columnwidth]{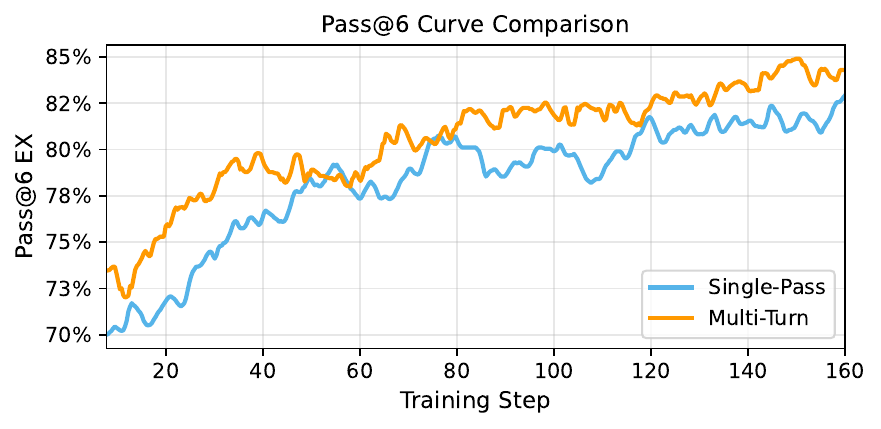}
    
    \caption{Training dynamics for Single-Pass vs. Multi-Turn RL. Curves represent the 10-step moving average of raw rewards (top) and Pass@6 accuracy (bottom).}
    \label{fig:single-multi}
\end{figure}
In this section, we analyze why multi-turn RL outperforms single-pass RL. We train both settings on the same RL dataset with identical hyperparameters and the same Qwen2.5-Coder-7B base model (no SFT), and evaluate majority-vote EX on Spider-dev/test and BIRD-dev (Table~\ref{tab:single_multi}). Because single-pass RL lacks interaction, we disable the turn-count reward in both settings and keep all other reward components fixed. Multi-turn RL consistently wins across benchmarks, with the largest gain on cross-domain BIRD-dev (+3pp, 56.3\%$\rightarrow$59.3\%), indicating stronger OOD generalization.

Figure~\ref{fig:single-multi} sheds light on the mechanism. Single-pass RL improves rapidly but saturates early, suggesting a lower capability ceiling. Multi-turn RL continues to scale, consistent with richer feedback from interactive execution and a longer optimization horizon. Notably, Pass@6 favors multi-turn RL throughout training, even when its reward curve is initially lower, implying that multi-turn training induces more diverse and exploratory trajectories rather than optimizing a single deterministic chain.

Qualitative examples in Appendix~\ref{app:single_multi} further support this: multi-turn agents handle ambiguous questions and complex schemas by iteratively revising hypotheses and verifying via execution, while single-pass RL lacks these self-correction behaviors. Together, these results explain the higher headroom, broader search, and stronger generalization of multi-turn RL.

\subsection{Post-training analysis}

In this section, we examine how post-training improves the base model through a series of extensive ablation studies. We analyze three key components: (1) the contribution of each reward in the RL stage, (2) how RL enhances the model’s reasoning ability, and (3) the effect of SFT as a cold-start initializer.

\subsubsection{Ablation study}
We perform a comprehensive reward and post-training stage ablation study on BIRD-dev in Appendix~\ref{app:ablation} Table~\ref{table:reward_ablation} to isolate the contribution of each component in our reinforcement learning framework. For SQL quality evaluation, we report average syntax accuracy, schema similarity, and bi-gram similarity, with calculation details the same as presented in Appendix~\ref{app:reward_panel}. To assess the multi-turn behavior of the agents, we also track the average number of conversation turns required to reach a solution. We compare the performance of our method against the Qwen2.5-Coder-7B base model, the post-SFT checkpoint, and Sonnet 3.7 as a large closed-source reference.

\paragraph{Reward ablation} The reward ablation results show that the bi-gram similarity reward provides the largest individual gain among partial rewards. Beyond encouraging surface overlap, bi-gram matching supplies a dense learning signal at the token-transition level, which reduces reward variance and steers decoding toward stable SQL skeletons early in optimization. In contrast, the execution reward is more sparse and less informative. As a result, improving bi-gram overlap—especially around high-impact transitions such as column references, JOIN–ON patterns, and GROUP BY/aggregation templates—more directly corrects the structural failure modes that dominate execution errors. Counterintuitively, format and execution rewards contribute only marginal gains once strong bi-gram, schema, and syntax constraints are present, suggesting substantial redundancy among these signals in the high-constraint regime.

\paragraph{Post-training ablation} Post-training ablations show scale-dependent behaviors. Larger models (e.g., Sonnet) are syntactically strong but weak at schema linking, and both Sonnet and base Qwen generate overly long trajectories. Without multi-turn post-training, models lack robust tool-use and multi-step planning: Sonnet wastes turns probing the schema, while base Qwen repeatedly revises flawed SQL due to weaker syntax. Comparing base-Qwen → SFT → RL highlights a clear progression: SFT improves syntax accuracy and shortens trajectories, but schema linking remains difficult; RL further improves schema identification and drives convergence to correct solutions within a tight turn range, indicating more efficient schema linking and reliable SQL generation. Appendix~\ref{app:ablation} reports matched-condition test-time compute comparisons (diversity vs. consistency) and a cold-start SFT ablation across scales.

\begin{figure}[h]
    \centering
    
    \includegraphics[width=\columnwidth]{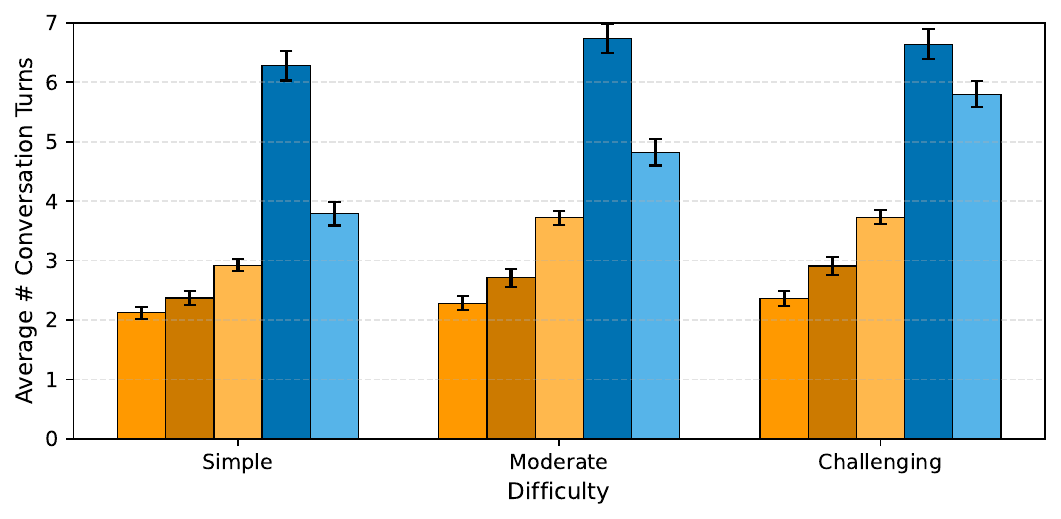}
    \vspace{-10pt}
    
    \includegraphics[width=\columnwidth]{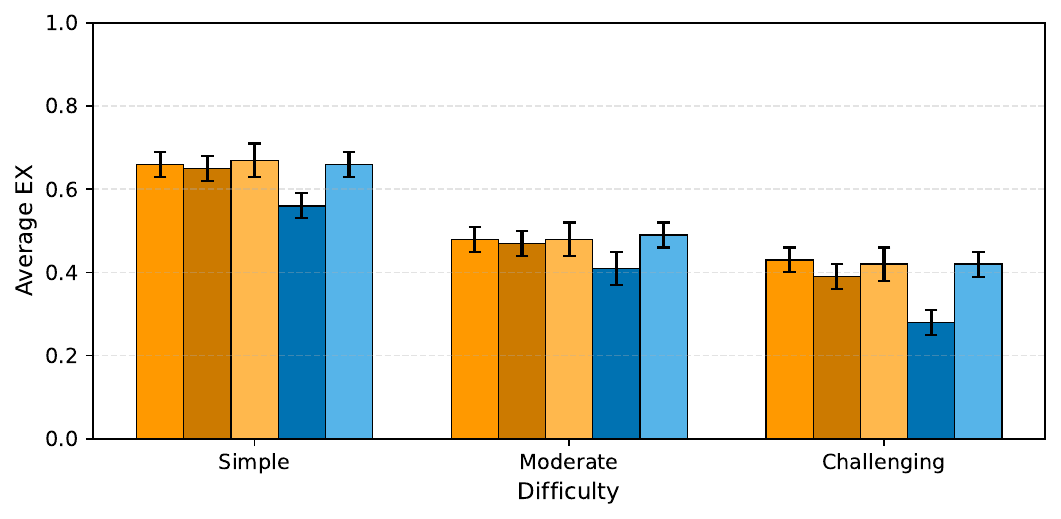}
    
    \caption{Performance comparison on the BIRD-dev set across three difficulty levels. The top panel reports the average number of turns, while the bottom panel displays the Average Execution Accuracy (EX). The models evaluated are 
\textcolor{trailRL}{\textbf{\method RL}}, 
\textcolor{trailSFT}{\textbf{\method SFT}}, 
\textcolor{trailNoReward}{\textbf{\method RL w/o turn reward}}, 
\textcolor{qwenAgent}{\textbf{Qwen2.5-Coder Agent}}, and 
\textcolor{sonnetAgent}{\textbf{Sonnet Agent}}.}
    \label{fig:turn}
\end{figure}

\paragraph{Turn-Efficiency}
We further investigate how RL and the turn-efficiency reward demonstrate a clear impact on interaction behavior. As shown in Appendix~\ref{app:ablation} Table~\ref{table:reward_ablation}, removing it causes the model to engage in longer, unnecessary reasoning chains without yielding higher accuracy, confirming that this signal suppresses over-thinking and encourages concise trajectories. Beyond simple turn reduction, Figure~\ref{fig:turn} shows that the reward enables adaptive turn budgeting: the agent takes more turns on harder queries and fewer on easier ones. It's also worth noticing that RL-trained \method shows advantages in solving challenging problems exceeding Sonnet-3.7 with significantly fewer turns. This difficulty-aware allocation improves execution accuracy by ensuring that the model expends effort only where needed, rather than uniformly across tasks. In Appendix~\ref{app:ablation} Figure~\ref{fig:single_multi_demo}, we present a side-by-side comparison where \method solves the problem efficiently, whereas Sonnet exhibits significant overthinking, yielding unnecessarily long trajectories that ultimately result in incorrect predictions.

\subsubsection{Reasoning analysis}
\begin{figure}
    \centering
    \includegraphics[width=1\linewidth]{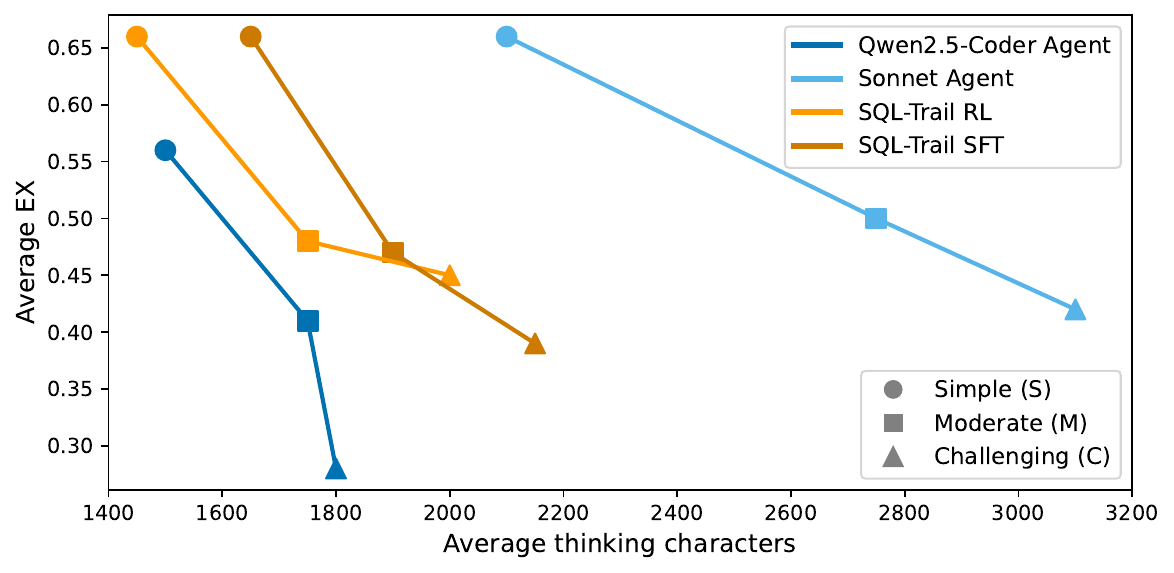}
    \caption{Reasoning efficiency analysis on BIRD-dev. We plot Execution Accuracy against the average quantity of reasoning (measured in characters within \think  tags) across different difficulty levels}
    \label{fig:reasoning-efficiency}
\end{figure}
To assess reasoning efficiency, we analyze the relationship between thinking-token length and execution accuracy in Figure~\ref{fig:reasoning-efficiency}. As expected, harder questions generally require longer reasoning traces to recover the correct SQL logic. However, base models fail to navigate this trade-off: Sonnet tends to overthink with diminishing returns, while Qwen-base frequently underthinks. In contrast, \method achieves balanced reasoning efficiency—reaching the highest execution accuracy with substantially fewer reasoning tokens, especially after RL optimization. This conclusion is further supported by our analysis of reasoning length versus schema-link complexity in Appendix~\ref{app:ablation} Figure~\ref{fig:reason_schema}. We further include analysis of reasoning efficiency evolution in Appendix~\ref{app:ablation} Figure~\ref{fig:training_dynamics}.

%% file: tables/main_table.tex
\begin{table*}[h]
  \centering
  \resizebox{\linewidth}{!}{
  \begin{tabular}{l c ccccccc c c}
    \hline
    \textbf{Text2SQL Method} &
    \textbf{Training Set (Size)} &
    \multicolumn{2}{c}{\textbf{Spider (dev)}} &
    \multicolumn{2}{c}{\textbf{Spider (test)}} &
    \multicolumn{2}{c}{\textbf{BIRD (dev)}} &
    \multicolumn{2}{c}{\textbf{Data Efficiency (Gre)$\uparrow$}} \\
    &
    &
    Gre & Maj &
    Gre & Maj &
    Gre & Maj &
    Spider-test & BIRD-dev \\
    \hline

    \multicolumn{10}{c}{\textbf{Sonnet-3.7}} \\
    \hline
    Single-pass & - & 78.3 & 78.9 & 82.0 & 83.2 & 58.5 & 60.1 & - & - \\
    Multi-turn Agent & - & 77.2 & 77.9 & 81.9 & 82.0 & 60.0 & 60.8 & - & - \\
    \hline

    \multicolumn{10}{c}{\textbf{Qwen2.5-Coder-3B-Instruct}} \\
    \hline
    Multi-turn Agent & - & 72.8 & 77.0 & 75.1 & 77.2 & 45.2 & 50.5 & - & - \\
    SQL-R1-3B & SynSQL(5k) & 71.9 & 78.1 & 76.5 & 78.9 & 48.4 & 54.6 & 0.28 & 0.64 \\
    \method-3B & SynSQL(0.8k)+Spider(1k) &
      \textbf{76.3} & \textbf{83.1} &
      \textbf{77.7} & \textbf{84.3} &
      \textbf{50.1} & \textbf{55.1} &
      \textbf{1.3} & \textbf{2.4} \\
    \hline

    \multicolumn{10}{c}{\textbf{Qwen2.5-Coder-7B-Instruct}} \\
    \hline
    Multi-turn Agent & - & 73.4 & 77.1 & 82.2 & 85.6 & 50.9 & 61.3 & - & - \\
    Reasoning-SQL-7B & BIRD-train(8026) & - & - & 78.7 & - & \textbf{64.0} & - & -0.44 & 1.63 \\
    SQL-R1-7B & SynSQL(5k) & 81.9 & 84.5 & 83.5 & 86.1 & 58.9 & 63.1 & 0.26 & 1.6 \\
    OminiSQL-7B & SynSQL(2.5M)+BIRD(9.4k)+Spider(8.7k) &
      81.2 & 81.6 & \textbf{87.9} & 88.9 & 63.9 & \textbf{66.1} & 0.002 & 0.005 \\
    \method-7B & SynSQL(0.8k)+Spider(1k) &
      \textbf{85.2} & \textbf{86.8} &
      86.0 & \textbf{87.6} &
      60.1 & 64.2 &
      \textbf{1.90} & \textbf{4.60} \\
    \hline

    \multicolumn{10}{c}{\textbf{Qwen2.5-Coder-14B-Instruct}} \\
    \hline
    Multi-turn Agent & - & 78.1 & 80.6 & 86.6 & 88.0 & 61.5 & 66.1 & - & - \\
    Reasoning-SQL-14B & BIRD-train(8k) & - & - & 81.4 & - & \textbf{65.3} & - & -0.1 & 0.34 \\
    SQL-R1-14B & SynSQL(5k) & 83.4 & 86.7 & 86.1 & 88.1 & 63.2 & \textbf{67.1} & -0.65 & 0.47 \\
    Omini-SQL-14B & SynSQL(2.5M)+BIRD(9.4k)+Spider(8.7k) &
      81.4 & 82.0 & \textbf{88.3} & 88.3 & 64.2 & 65.9 & 0.0006 & 0.001 \\
    \method-14B & SynSQL(0.8k)+Spider(1k) &
      \textbf{85.1} & \textbf{87.1} &
      86.8 & \textbf{88.5} &
      63.6 & 66.7 &
      \textbf{0.1} & \textbf{1.05} \\
    \hline
  \end{tabular}}

  \caption{\label{tab:main_results}
Main EX(\%) and data efficiency results. The table is organized into blocks, each headed by the corresponding base model. "Multi-turn Agent" refers to untuned base models initiated with a multi-turn system template; see more details in Appendix~\ref{app:systemprompt}. “Gre” denotes greedy decoding and “Maj” denotes majority voting.}

\end{table*}

%% file: tables/single_vs_multi.tex
\begin{table}
\resizebox{\columnwidth}{!}{
  \centering
  \begin{tabular}{lccc}
    \hline
    \textbf{Model} &
    \textbf{Spider (Dev)} &
    \textbf{Spider (Test)} &
    \textbf{BIRD (Dev)} \\
    \hline
    single-pass RL & 82.8 & 85.1 & 56.3 \\
    multi-turn RL  & 84.5 & 86.1 & 59.3 \\
    \hline
  \end{tabular}}
  \caption{\label{tab:single_multi}
Execution accuracy(\%) comparison between single-turn RL and multi-turn RL under identical training configurations. Results are reported using majority voting.}
\end{table}

%% file: sections/conclusion.tex
\section{Conclusion} We introduce \method, a unified multi-turn RL framework that reframes Text-to-SQL from static translation into interactive reasoning. Using difficulty-aware rewards and targeted data selection, \method enables open-source models to explore schemas, correct errors via execution feedback, and adapt their turn budget to query complexity. \method sets a new state of the art in data efficiency and out-of-distribution generalization, matching or outperforming larger proprietary systems on BIRD-dev with fewer than 2,000 training samples. These results highlight the importance of iterative environment interaction for robust, self-correcting database agents.

%% file: sections/Appendix.tex
\subsection{Potential Risks}
Deploying multi-turn Text-to-SQL agents introduces privacy, security, and operational risks. Because \method explicitly queries the database during inference, it may surface sensitive information via returned rows, error messages, or intermediate tool traces. This risk is amplified by multi-turn exploration, which can adaptively probe schemas and values. To mitigate this, deployments should enforce least-privilege access (ideally read-only), redact or aggregate sensitive outputs, and log/access-audit queries and results with appropriate retention policies.

There are also security and misuse risks: adversarial schema/table names or database contents can act as prompt-injection vectors that steer the agent toward unsafe behavior, and the agent could generate destructive or exfiltrative SQL if permissions allow (e.g., \texttt{UPDATE}/\texttt{DROP}, wide table scans). Practical safeguards include strict SQL allow-lists (e.g., \texttt{SELECT}-only), static query analysis, query cost estimation, timeouts, rate limits, and sandboxed execution. Operationally, multi-turn agents can increase database load and may inadvertently issue expensive queries, creating denial-of-service-style failures under concurrency. Finally, as with many LLM systems, performance may vary across domains and languages underrepresented in training/evaluation, potentially creating unequal reliability across user groups; careful domain-specific evaluation and monitoring are required before high-stakes use.

\subsection{Dataset statistics}
Table~\ref{tab:dataset_splits} summarizes the dataset sizes used in this work. Spider and BIRD provide standard train/dev/test splits for supervised training and evaluation, with BIRD’s test split typically kept hidden for leaderboard evaluation. SynSQL-2.5M is a large-scale synthetic corpus released primarily as a training set, which we use for training but do not treat as a canonical benchmark dev/test split.
\begin{table}[t]
\centering
\small
\begin{tabular}{lrrrr}
\toprule
\textbf{Dataset} & \textbf{Train} & \textbf{Dev} & \textbf{Test} & \textbf{Total} \\
\midrule
Spider & 7{,}000 & 1{,}034 & 2{,}147 & 10{,}181 \\
BIRD & 9{,}428 & 1{,}534 & 1{,}789 & 12{,}751 \\
SynSQL-2.5M & 2{,}536{,}035 & -- & -- & 2{,}536{,}035 \\
\bottomrule
\end{tabular}
\caption{Dataset sizes (question--SQL pairs). SynSQL-2.5M is released primarily as a large synthetic training corpus in the public release we use, without a canonical dev/test split.}
\label{tab:dataset_splits}
\end{table}

\subsection{Multi-turn agentic system design.}\label{app:agent_workflow}
Our approach follows an iterative framework where LLM can alternate between natural-language reasoning and external SQL execution in a closed loop~\citep{cao2025skyrl}. The system instruction enforces a strict interface: at each intermediate turn, after receiving the current environment input, the model must begin its response with a block of reasoning enclosed in \think{} and \thinkend{}. It must then place its proposed SQL query for that step between \sql{} and \sqlend{} at the end of the response. Once these tokens appear, the system extracts the enclosed SQL and forwards it to the SQL engine. The resulting execution output is wrapped between \obs{} and \obsend{} and appended to the conversation as user input for the next iteration. At every turn, the LLM conditions on the full history of past actions and observations to produce its next action, consisting of a new reasoning trace and SQL query. This loop continues until either a maximum turn limit is reached or the model outputs a final answer by enclosing its completed SQL solution between \sol{} and \solend{}. We guide the initial LLM to follow our predefined instructions using a detailed system prompt template, as shown in Appendix~\ref{app:agent_workflow}. This generation procedure serves as the central inference scaffold for both rollout during training and evaluation. By enforcing strict formatting rules, it constrains the model’s behavior and ensures accurate reward assignment during the RL training stage.

As shown in System Prompt~\ref{app:systemprompt}, we explicitly instruct the LLM to first identify all relevant tables and columns from the provided database schema before issuing any SQL query. In particular, the model must list the tables and fields it plans to use in its initial <think> block, ensuring that subsequent reasoning and tool calls are grounded in the schema rather than hallucinated structures. The prompt further enforces that the final \texttt{SELECT} clause only includes columns explicitly requested in the natural language question, preventing over-selection and spurious attributes. Together, these constraints encourage disciplined schema selection and promote faithful, schema-aware SQL generation throughout the multi-turn interaction.

We provide our multi-turn generation workflow in Algorithm~\ref{alg:multiturn-sql}. The procedure iteratively rolls out the policy’s token-level decisions, alternates between model reasoning and SQL tool calls, and injects execution feedback back into the dialogue history to guide subsequent turns. Notably, when returning SQL execution results, we deliberately serialize data frames with column headers included, ensuring the agent receives richer schema context during database probing and enabling more reliable schema linking in later turns.

During the prompt and workflow optimization stage, we evaluate our prompt-design choices using Sonnet-3.7 on BIRD-dev (Table~\ref{tab:agent_prompt_design}). Our ablation progressively introduces database schema details, column headers in SQL execution observations, and explicit schema-selection instructions in the system prompt. The results highlight two key findings: (i) providing rich schema context through interactive database exploration substantially improves grounding, and (ii) enforcing schema-selection constraints keeps the model’s choices consistent throughout the trajectory.
\begin{table}[h]
\resizebox{\columnwidth}{!}{
  \centering
  \begin{tabular}{lc}
    \hline
    \textbf{Agent prompt design} &
    \textbf{BIRD (Dev)} \\
    \hline
    w/ Schema + DFCols + Selection guiding & 59.7 \\
    w/ Schema + DFCols                     & 58.8 \\
    w/ Schema                             & 55.9 \\
    w/o Schema                          & 28.4 \\
    \hline
  \end{tabular}}
  \caption{\label{tab:agent_prompt_design}
Execution accuracy (\%) on BIRD-dev under different agent prompt designs. “Schema” indicates whether the full database schema is included in the initial user prompt. “DFCols” denotes whether column headers are shown in the SQL execution return. “Selection guiding” refers to adding instructions in the system prompt to enforce strict selection of relevant tables and columns.}
\end{table}

\subsection{Detailed Reward Formulations} \label{app:reward_panel}

In this section, we provide the specific definitions and calculation methods for the four reward terms introduced in the main text.

\paragraph{Final execution reward}

The primary objective of the text-to-SQL task is to generate a query that retrieves the correct answer from the database. To reflect this, we introduce execution reward ($r_{\text{exec}}$). For each generated trajectory, we extract the final SQL solution from \sol{}...\solend{} tokens, and then we execute this query and compare the generated results with the gold SQL query execution results. If they are identical, we assign 1 to the execution reward term. This is the formal definition of it:
\begin{equation}
    r_{\text{exec}} = \mathbbm{1}[\text{exec}(\text{pred\_sql}) = \text{exec}(\text{gold\_sql})]
\end{equation}
\paragraph{Turn number reward}
The length of conversation is critical to long-horizon multi-turn agent behavior. To encourage efficiency and discourage the agent from engaging in redundant reasoning steps, we introduce a turn-based reward ($r_{\text{turns}}$). This term penalizes the agent for exceeding predetermined turn budget $T$ . The reward is conditional on staying within the budget and, for harder queries, achieving correct execution. The formulation is given by:
\begin{equation}
    r_{\text{turns}} = 
    \begin{cases} 
    1, & d = \text{simple} \land t \le 2 \\
    1, & d = \text{medium} \land t \le 3 \\
    1, & d \in \{\text{hard, extra}\} \land r_{\text{exec}} = 1 \land t < T \\
    0, & \text{otherwise}
    \end{cases}
\end{equation}
where $t$ is the finishing turn number of the current trajectory.

\paragraph{Syntax Correctness Reward}
A fundamental prerequisite for any generated code is that it must be executable by the database engine. To provide an early learning signal that distinguishes between valid SQL queries and hallucinated strings that violate SQL grammar, we utilize a syntax correctness reward $r_{\text{syntax}}$. This is a binary indicator function that returns 1 if the predicted SQL is valid and executable, and 0 otherwise:
\begin{equation}
    r_{\text{syntax}} = \mathbbm{1}[\text{pred\_sql is executable}]
\end{equation}

\paragraph{N-gram Similarity Reward}
Binary execution rewards are often too sparse as a generated query might be semantically close to the solution but fail to execute due to minor token mismatches. To provide a denser training signal, we incorporate an N-gram similarity reward $r_{\text{ngram}}$. This metric measures the lexical overlap between the bigrams ($n=2$) of the predicted query set $B_{\text{pred}}$ and the gold query set $B_{\text{gold}}$ using Jaccard similarity:
\begin{equation}
    r_{\text{ngram}} = \frac{|B_{\text{pred}} \cap B_{\text{gold}}|}{|B_{\text{pred}} \cup B_{\text{gold}}|}
\end{equation}
For example, consider a gold query \texttt{SELECT name FROM student} and a predicted query \texttt{SELECT name FROM teacher}. The set of bigrams for the gold query is \{\texttt{SELECT name}, \texttt{name FROM}, \texttt{FROM student}\}, while the prediction yields \{\texttt{SELECT name}, \texttt{name FROM}, \texttt{FROM teacher}\}. The intersection contains two bigrams and the union contains four distinct bigrams, resulting in a reward score of $2/4 = 0.5$.

\paragraph{Schema Linking Reward}
A common failure mode in text-to-SQL generation is the hallucination of table or column names. To enforce strict grounding, we calculate a schema linking reward $r_{\text{schema}}$ based on the Jaccard similarity between the set of schema items (tables and columns) appearing in the prediction $S_{\text{pred}}$ and the gold label $S_{\text{gold}}$:
\begin{equation}
    r_{\text{schema}} = \frac{|S_{\text{pred}} \cap S_{\text{gold}}|}{|S_{\text{pred}} \cup S_{\text{gold}}|}
\end{equation}
For instance, if the gold query retrieves data from the table \texttt{Employees} and column \texttt{Salary}, the gold schema set is \{\texttt{Employees}, \texttt{Salary}\}. If the model correctly identifies \texttt{Employees} but hallucinates a column \texttt{Wages}, the predicted set is \{\texttt{Employees}, \texttt{Wages}\}. The intersection is \{\texttt{Employees}\} and the union is \{\texttt{Employees}, \texttt{Salary}, \texttt{Wages}\}, yielding a reward of $1/3$.

\paragraph{Format Reward}
To ensure the model adheres to the formatting constraints specified in the system prompt, which is crucial for supporting the multi-turn agent workflow and downstream parsing, we grant reward only when the output contains well-formed tags. Specifically, the output must include \texttt{\textbackslash think} and \texttt{\textbackslash thinkend} to delimit the reasoning process, as well as \texttt{\textbackslash sol} and \texttt{\textbackslash solend} to encapsulate the final SQL solution:
\begin{equation}
    r_{\text{format}} = \mathbbm{1}[\text{final output has correct format}]
\end{equation}

\paragraph{Total Reward}
The final reward $R$ is a weighted sum:
\begin{equation}\small
\begin{aligned}
    Reward = &5 r_{\text{exec}} + 2 r_{\text{turns}}\\
    &+ r_{\text{schema}} + r_{\text{bigram}} + r_{\text{syntax}} + r_{\text{format}}  
\end{aligned}
\end{equation}
We set these weights empirically, with a significantly larger weight on execution since execution correctness is the primary objective and the most reliable learning signal for end-task performance. We assign a slightly larger weight to the turn-budget term to explicitly shape multi-turn agent behavior (encouraging concise yet sufficient interactions). All remaining auxiliary shaping terms are given unit weight, reflecting no strong prior preference among them; they serve mainly to stabilize training and provide additional guidance without dominating optimization.

\subsection{Training Configuraiton.}\label{app:train_confg}
For the SFT stage, we train with a batch size of 128 for two epochs and adopt an optimizer configuration consisting of a learning rate of $1\times10^{-5}$, betas of $(0.9, 0.95)$, a weight decay of 0.01, a warmup ratio of 0.1, gradient clipping at 1.0, and a cosine learning-rate schedule. For the RL stage, we initialize from the SFT checkpoint and continue training with the same batch size of 128, using a learning rate of $1\times10^{-6}$ and top-$p=0.99$ sampling for rollouts. Each question is expanded with a rollout group size of six under a maximum turn budget of ten, and we report evaluation results at step 108.

\subsection{Training data curation}\label{app:data_curation}
In our training pipeline, we applied a two-stage process: Supervised Fine-Tuning (SFT) followed by Reinforcement Learning (RL). To maximally utilize information from a relatively small dataset and investigate the data efficiency of multi-turn RL compared to other data-hungry fine-tuning methods, we deliberately filtered for informative training samples and carefully balanced their difficulty.

\paragraph{SFT} For supervised fine-tuning, we begin by randomly sampling 3,000 training examples from Spider-train~\citep{yu2018spider}. We then run inference with Claude-Sonnet-3.7~\citep{anthropic_claude_3.7_sonnet_system_card} using our multi-turn agent template from Section~\ref{sec:3.1}, allowing up to 10 turns per example. From these synthetic trajectories, we select 1,000 that produce correct final SQL predictions, prioritizing examples with moderate to high difficulty under Spider’s difficulty classification. These curated trajectories serve as high-quality demonstrations for SFT, enabling the student base models to distill strong instruction-following and multi-turn reasoning behaviors from a more capable closed-source model.

\paragraph{RL} For the RL stage, we constructed a dataset of 1,027 samples. To ensure data-effective learning and robust reasoning, we divided this data into two strategic categories: a difficulty-balanced set and an expanded exploration set.

The \textbf{Difficulty-balanced Set} (700 samples) is curated to prioritize "hard yet solvable" instances. As detailed in the GRPO formulation in Sec~\ref{sec:3.1}, the optimization process relies on the relative advantage within a group of sampled outputs. If a training input yields all correct or all incorrect responses across the sampling group, the reward variance is zero. Consequently, these samples provide no useful gradient signal, failing to utilize the dataset effectively. Furthermore, a high prevalence of such samples decreases the valid batch size, leading to training instability. To mitigate this, we sampled 2,800 candidates from the synthetic text-to-SQL corpus SynSQL-2.5M, prioritizing samples from challenging difficulty level, and evaluated the SFT model using $G=6$ stochastic generations per question. We computed a difficulty score, $\mathcal{S}(q)$, to identify items that are neither trivial nor impossible:
\begin{equation}
    \mathcal{S}(q) = 
    \begin{cases} 
    \text{pass}@G(q) & \text{if } 0 < \text{pass}@G(q) < 1 \\
    1 & \text{otherwise}
    \end{cases}
\end{equation}
By ranking samples in ascending order of $\mathcal{S}(q)$ and selecting the top 700, we focus on problems that are hard yet solvable where the model succeeds occasionally (e.g., $\text{pass}@6 \approx 1/6$) but does not solve them reliably. This filtering removes both overly easy and unsolvable cases, ensuring dense, informative advantage signals that improve optimization stability and data efficiency.

The \textbf{Exploration Set} (327 samples) targets challenging edge cases that encourage exploration and help correct persistent error modes. It is constructed by combining three sources of difficult instances: Post-SFT failures—127 questions from the original SFT training set that the SFT model still fails to solve; a subset of 100 challenging SynSQL examples with pass@6 = 0; and an additional 100 extra-hard Spider-train questions with pass@6 = 0. Together, these samples capture diverse but reliably difficult behaviors that the model must learn to overcome during RL training.

\subsection{Multi-turn v.s. Single-pass}\label{app:single_multi}
\input{tables/base_model_comp}
To further evaluate generalization under distribution shift, we assess \method on three robustness benchmarks: \textbf{Spider-DK}, which injects domain knowledge into a subset of databases and stresses joint schema--knowledge reasoning; \textbf{Spider-Syn}, which rewrites questions with synonyms to test robustness to lexical variation; and \textbf{Spider-Realistic}, which removes schema-name leakage to approximate more natural user phrasing. We compare against strong single-pass baselines, including SQL-R1 and Reasoning-SQL---both trained with single-pass GRPO---as well as OminiSQL, which relies on single-pass SFT.

As shown in Table~\ref{table:robust}, \method consistently matches or surpasses all single-pass baselines across all robustness settings. The largest gains arise on \textbf{Spider-Syn}, where the multi-turn formulation improves accuracy by a substantial margin, indicating superior resilience to lexical perturbations. Notably, \method also achieves the best performance on \textbf{Spider-Realistic}, demonstrating strong grounding when explicit schema cues are removed. These results collectively highlight that multi-turn RL provides intrinsic advantages over single-pass formulations, particularly when the model must reason over noisy, perturbed, or partially obscured database schemas.

\begin{figure}[t]
    \centering
    
    \includegraphics[width=1.1\linewidth]{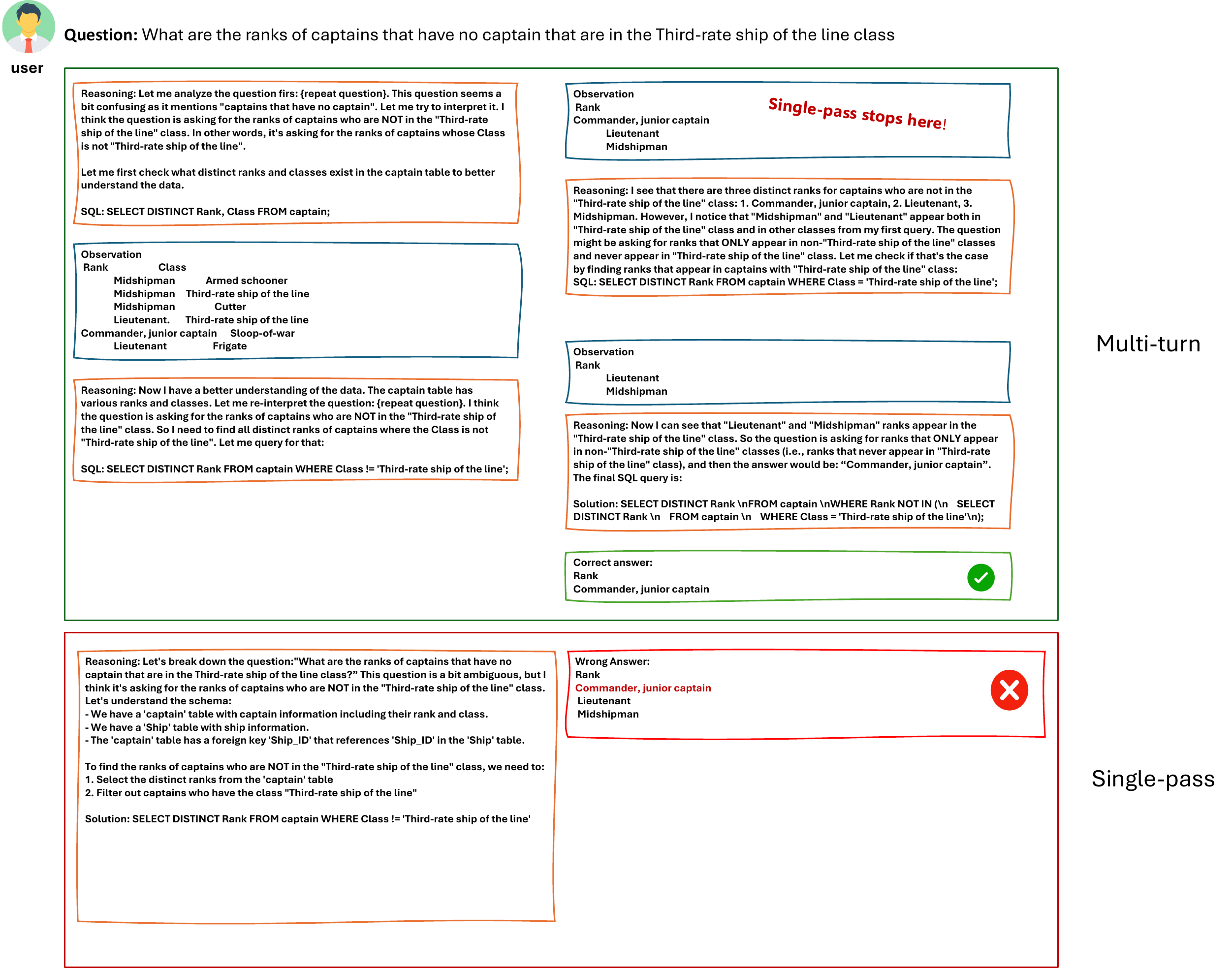}
    
    
    \caption{Example where multi-turn succeeds single-pass}
    \label{fig:single_multi_demo}
\end{figure}

As shown in Figure~\ref{fig:single_multi_demo}, we provide side-by-side examples of a single-pass LLM and a multi-turn agent. The qualitative comparison illustrates that multi-turn agents handle difficult questions and noisy or ambiguous databases far more effectively. Notably, in the multi-turn trajectory, the agent initially makes the same mistake as the single-pass model on the second turn. However, instead of committing to this incorrect answer, it reflects on the question, conducts additional probing, and carefully re-examines the schema. This iterative refinement allows the multi-turn agent to correct its earlier error and ultimately arrive at the correct solution—behavior that single-pass LLMs are unable to exhibit.

\subsection{Additional experiments}\label{app:ablation}
\input{tables/ablation}
\input{tables/cold_start}

\begin{figure}
    \centering
    \includegraphics[width=1\linewidth]{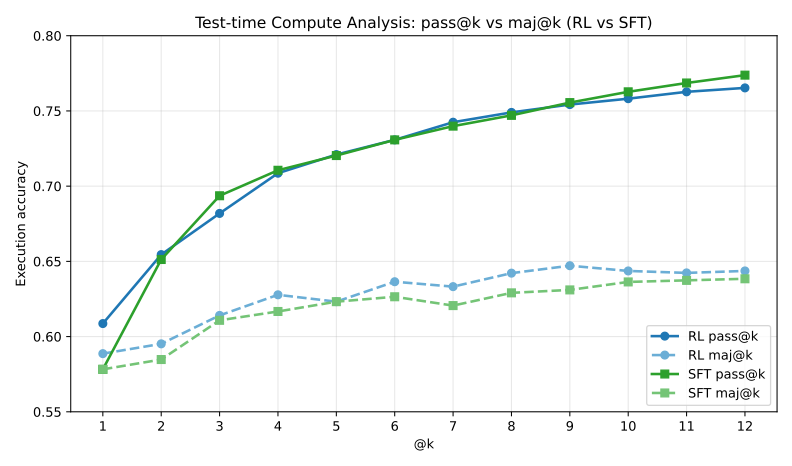}
    \caption{Test-time compute analysis}
    \label{fig:test-time}
\end{figure}

\paragraph{Test-time compute.}
We analyze the impact of test-time compute scaling in Figure~\ref{fig:test-time}, comparing the SFT baseline against the RL-tuned model. As the number of samples $k$ increases, the SFT model eventually overtakes the RL model on Pass@$k$, indicating that supervised fine-tuning preserves greater sample diversity. Conversely, the RL model consistently outperforms SFT on Majority@$k$, suggesting that reinforcement learning encourages the model to converge on correct solutions with higher consistency, albeit at the cost of diversity. 
\begin{figure}
    \centering
    \includegraphics[width=1\linewidth]{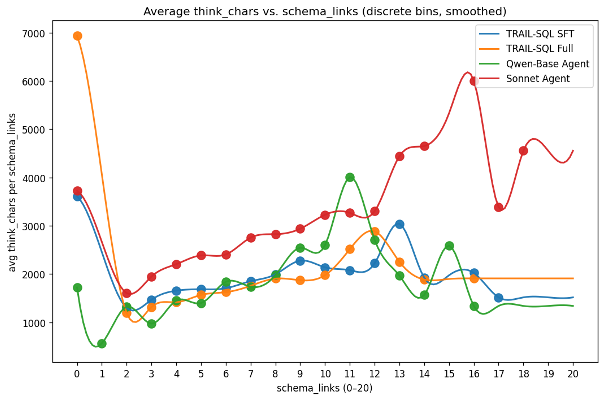}
    \caption{Reasoning efficiency v.s schema links}
    \label{fig:reason_schema}
\end{figure}
\paragraph{Cold-start ablation} Additionally, we examine the efficacy of initialization strategies in Table~\ref{tab:cold_start}. The results demonstrate that RL with SFT cold-start yields the highest performance, surpassing both the RL model trained without cold-start and the SFT-only baseline.
\begin{figure}
    \centering
    \includegraphics[width=1\linewidth]{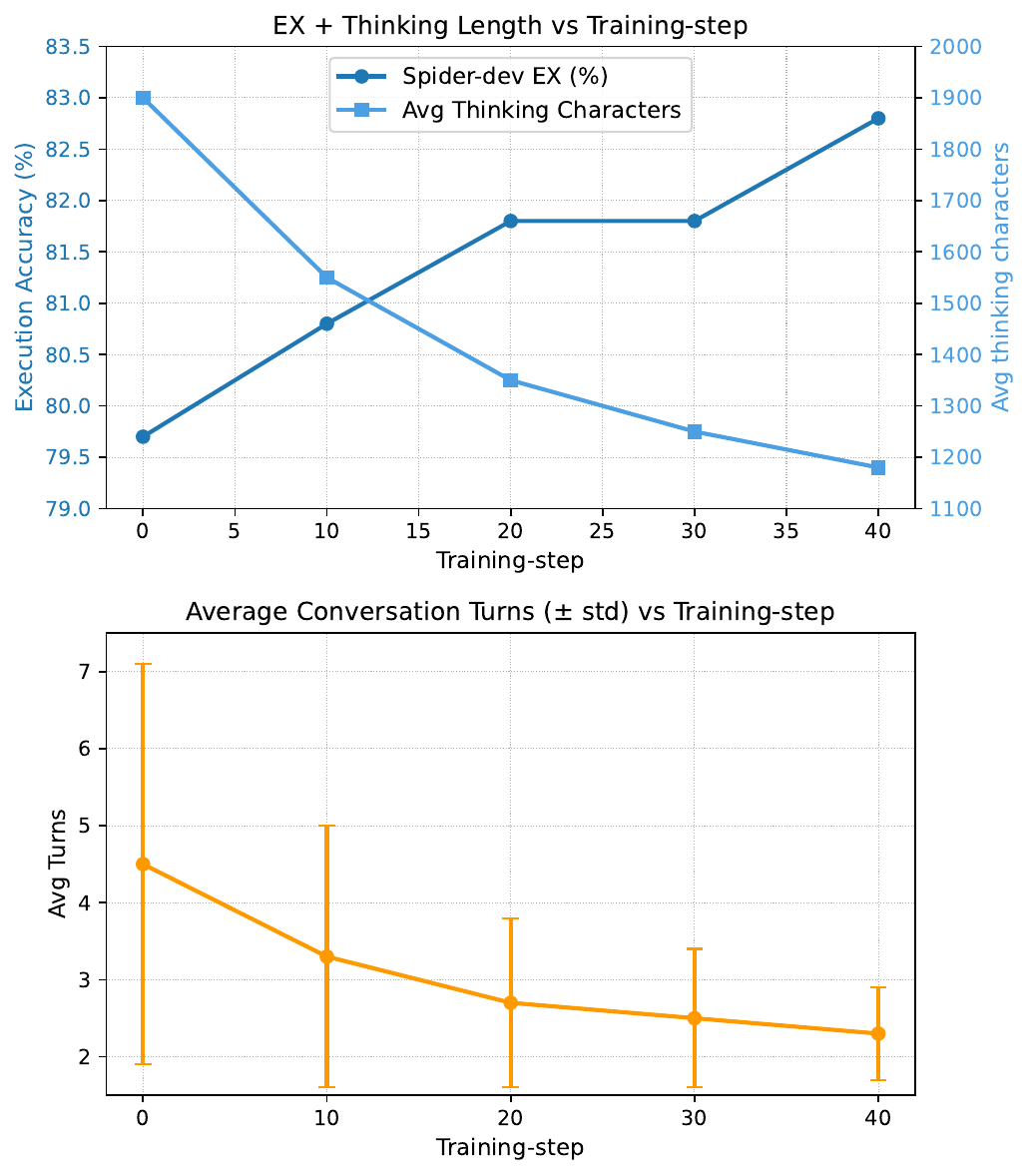}
    \caption{Changes in reasoning efficiency during training, shown alongside the curves for execution accuracy and the number of conversation turns.}
    \label{fig:training_dynamics}
\end{figure}
\paragraph{Reasoning scaling with schema complexity}
Figure~\ref{fig:reason_schema} characterizes the relationship between reasoning length (average characters within \texttt{<think>} tags) and the number of schema links in the predicted SQL. Generally, a higher density of schema links necessitates increased reasoning volume. However, we observe distinct resource allocation patterns across models: Sonnet tends to generate excessive reasoning chains, whereas Qwen-base often exhibits insufficient reasoning depth. In contrast, \method demonstrates an adaptive allocation, scaling its reasoning effort effectively to resolve schema dependencies without incurring unnecessary computational overhead.

\paragraph{Evolution of reasoning efficiency}
We further analyze the training dynamics by monitoring execution accuracy alongside generation costs in Figure~\ref{fig:training_dynamics}. As training progresses, we observe that execution accuracy steadily improves, while both the reasoning character length and the total trajectory length decrease. This inverse correlation suggests that the model learns to optimize its thought process, pruning redundant reasoning steps to achieve correct solutions with greater computational efficiency.

\onecolumn

\begin{figure}[t]
    \centering
    
    \includegraphics[width=1.1\linewidth]{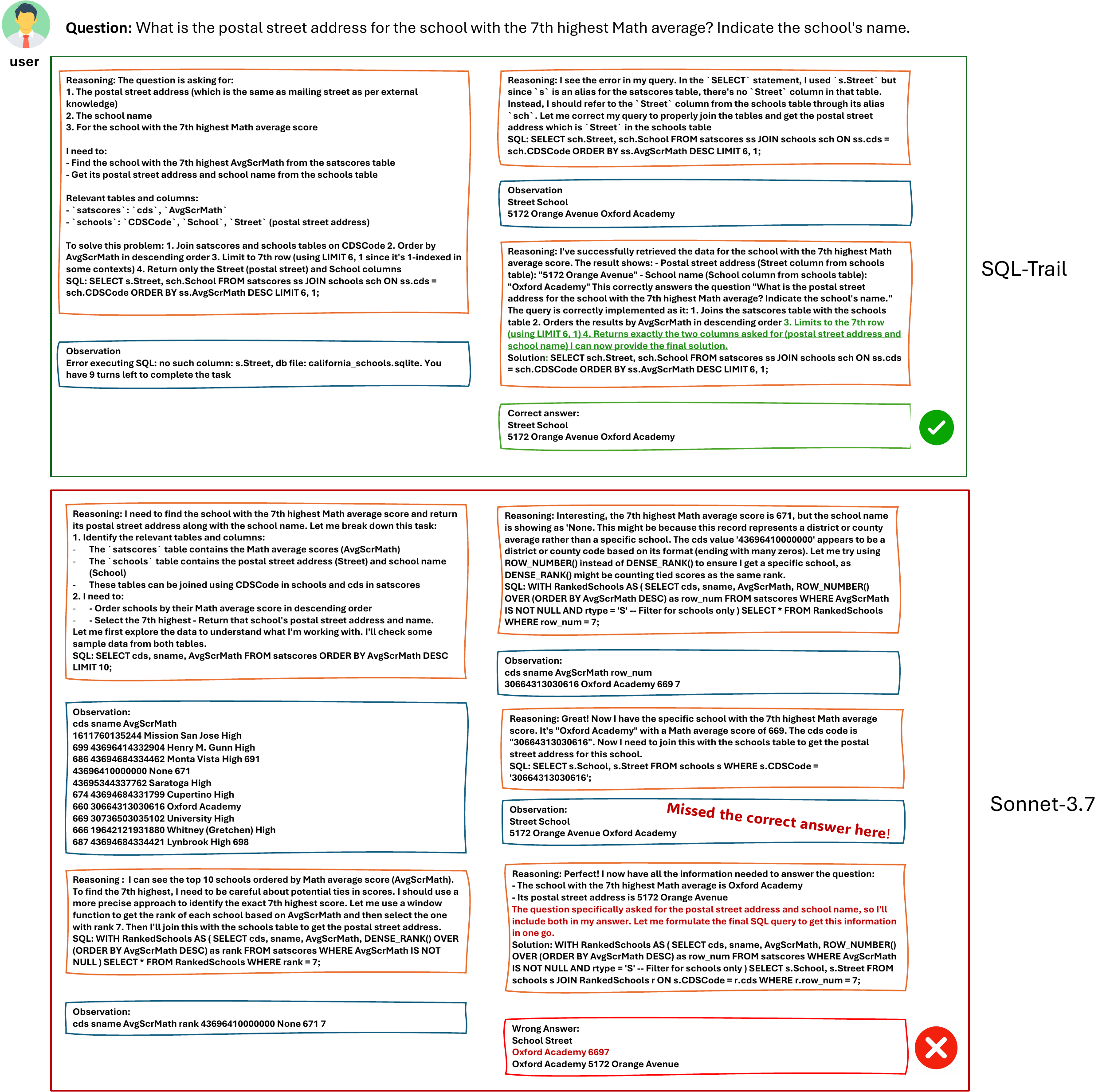}
    
    
    \caption{\method and Sonnet-3.7 agent overthinking behavior comparison}
    \label{fig:overthink}
\end{figure}

\begin{algorithm}[h]
\caption{LLM Response Rollout with Multi-Turn SQL Tool Calls}
\label{alg:multiturn-sql}
\begin{algorithmic}[1]
\Require Input question $x$, policy model $\pi_\theta$, maximum turn budget $T$
\Ensure Final SQL query $q^\star$
\State Initialize dialogue history $h \gets x$
\State Initialize turn counter $t \gets 0$
\While{$t < T$}
    \State Initialize current turn response $y_t \gets \varnothing$
    \While{True}
        \State Generate next token $u \sim \pi_\theta(\cdot \mid x, h, y_t)$
        \State Append token $y_t \gets y_t + u$
        \If{$u \in [\thinkend, \sqlend, \solend, \texttt{<eos>}]$}
            \State \textbf{break}
        \EndIf
    \EndWhile
    \State Append assistant turn $h \gets h + y_t$
    \If{\sol detected in $y_t$}
        \State Extract final SQL $q^\star$ from \sol block
        \State \Return $q^\star$
    \ElsIf{\sql detected in $y_t$}
        \State Parse SQL query $\hat{q} \gets \textsc{ParseSQL}(y_t,\sql)$
        \State Execute query to obtain result $r \gets \textsc{ExecSQL}(\hat{q})$
        \If{$r$ is invalid}
            \State $o \gets$ ``Your previous action is invalid. Think and try again.''
        \Else
            \State $o \gets \textsc{ToString}(\textsc{DataFrame}(r)\text{ with column headers})$
        \EndIf
        \State
        $z \gets \obs~ o +
        \text{``You have $(T-t)$ turns left to complete the task.``} \obsend$
        \State Append user observation $h \gets h + z$
    \EndIf
    \State $t \gets t + 1$
\EndWhile
\State Generate final \sol...\solend  block $y^{\text{sol}} \sim \pi_\theta(\cdot \mid x, h)$
\State Extract and return final SQL $q^\star$ from $y^{\text{sol}}$
\end{algorithmic}
\end{algorithm}

\onecolumn
\centering
\begin{PromptBox}[title=System Prompt for Multi-turn Text-to-SQL Agent]\label{app:systemprompt}
Task Overview:
You are a helpful text-to-SQL expert assistant. Below, you are provided with a database schema and a natural language question. Your task is to understand the schema and generate a valid SQL query
to answer the question within limited turns. You should
breakdown the problem, draft your reasoning process, and
generate the solution. \\

Database Engine: \{engine\}\\

Database Schema: \{schema\}\\

External Knowledge: \{external\_knowledge\}\\

Question: \{question\}\\

Instructions:
\begin{enumerate}
    \item Make sure you only output the information that is asked in
the question. If the question asks for a specific column,
make sure to only include that column in the SELECT clause,
nothing more.
    \item The generated query should return all of the information
asked in the question without any missing or extra
information.
\item In your first response, you must identify all tables and columns from the database schema that are relevant to answering the question.
    \item Before generating the final SQL query you should reflect through previous steps and refine any errors
previous steps.
\end{enumerate}

Format:
\begin{enumerate}
    \item Conduct thinking inside <reasoning>...</reasoning> blocks every
time you get new observation or information.
nothing more.
    \item You can use SQL tool written within a single <sql>your
sql</sql> block to explore or verify. SQL tool output will
be shown as dataframe inside <observation>...</observation>.
Based on this observation, you can think again and refine.
    \item The returned dataframe will be truncated in 50 rows if
observation is too long.
    \item If you find no further exploration is needed or reaches max turns, directly provide the final sql in <solution>…</solution>
\end{enumerate}
\begin{verbatim}
------------------------ START OF EXAMPLE ------------------------
Question: what is the average price of all products?

Database Schema:
    Table: products
    - id (INTEGER, PRIMARY KEY)
    - name (TEXT)
    - price (REAL)

<think>I need to find the average price of all items listed in the products table. 

Relavent tables and columns:
- 'products':'price'

To solve this problem I should issue a simple aggregate query using 
AVG(price).</think>
<sql>SELECT AVG(price) FROM products;</sql>
<observation>
   AVG(price)
0       24.75
</observation>
<think>The result shows that the average product price is 24.75. The question asks 
for the average price, so I can directly return this SQL query as the final 
solution.</think>
<solution>SELECT AVG(price) FROM products;</solution>
------------------------ END OF EXAMPLE ------------------------
\end{verbatim}
\end{PromptBox}

%% file: tables/base_model_comp.tex
\begin{table*}[ht]
  \centering
  \begin{tabular}{cl|cc|cc|cc}
    \hline
    \multicolumn{2}{c|}{\textbf{Setup}} &
    \multicolumn{2}{c|}{\textbf{Spider-DK}} &
    \multicolumn{2}{c|}{\textbf{Spider-Syn}} &
    \multicolumn{2}{c}{\textbf{Spider-Realistic}} \\
    \textbf{Base LLM} & \textbf{Method} &
    \textbf{Gre} & \textbf{Maj} &
    \textbf{Gre} & \textbf{Maj} &
    \textbf{Gre} & \textbf{Maj} \\
    \hline
    \multirow{5}{*}{Qwen2.5-Coder-7B} &
      Multi-turn Agent & 67.5 & 73.6 & 63.1 & 66.9 & 66.7 & 70.5 \\
    & SQL-R1 & -- & \textbf{78.1} & -- & 76.7 & -- & 83.3 \\
    & Reasoning-SQL & 73.3 & -- & 69.3 & -- & -- & -- \\
    & Omini-SQL & 76.1 & 77.8 & 69.7 & 69.6 & 76.2 & 78.0 \\
    & \method &
      \textbf{76.8} & \textbf{78.1} &
      \textbf{72.8} & \textbf{77.0} &
      \textbf{79.6} & \textbf{83.9} \\
    \hline
  \end{tabular}
  \caption{\small
    Robustness Benchmark experiment. "Multi-turn Agent" means non-finetuned base LLM initiated with multi-turn system prompt.
  }
  \label{table:robust}
\end{table*}


%% file: tables/ablation.tex
\begin{table*}
  \centering
  \resizebox{\linewidth}{!}{
  \begin{tabular}{cl|ccc|cc|cc}
    \hline
    \multicolumn{2}{c|}{\textbf{Setup}} &
    \multicolumn{3}{c|}{\textbf{Quality Metrics (\%)}} &
    \multicolumn{2}{c|}{\textbf{Conversation Turns}} &
    \multicolumn{2}{c}{\textbf{BIRD-dev EX (\%)}} \\
    \textbf{Model} & \textbf{Reward} &
    \textbf{Syntax Acc}$\uparrow$ &
    \textbf{Schema Sim}$\uparrow$ &
    \textbf{Bi-gram Sim}$\uparrow$ &
    \textbf{Avg} & \textbf{Std} &
    \textbf{Gre}$\uparrow$ & \textbf{Maj}$\uparrow$ \\
    \hline

    Multi-turn Sonnet 3.7  & -- &
      \textbf{98.6} & 81.9 & 63.3 & 4.24 & 2.22 & 60.0 & 60.8 \\
    \hline

    Multi-turn Qwen2.5-Coder-7B & -- &
      89.6 & 78.3 & \textbf{70.3} & 6.44 & 3.81 & 49.1 & 51.2 \\
    \hline

    Post-SFT \method-7B & -- &
      97.8 & 85.1 & 66.7 & 2.53 & 1.51 & 57.8 & 58.7 \\
    \hline

    \multirow{7}{*}{\method-7B} &
      all rewards & 97.8 & \textbf{86.2} & 67.7 & 2.26 & 0.65 & \textbf{60.1} & \textbf{64.2} \\
    & - w/o turns        & 97.4 & 85.5 & 67.5 & 3.19 & 2.12 & 59.3 & 63.2 \\
    & - w/o ngram        & 97.3 & 85.1 & 66.9 & 2.35 & 0.97 & 57.2 & 61.6 \\
    & - w/o schema       & 96.4 & 84.7 & 66.5 & 2.28 & 0.69 & 58.5 & 62.6 \\
    & - w/o syntax       & 96.9 & 85.7 & 67.5 & 2.28 & 0.72 & 59.1 & 63.8 \\
    & - w/o format       & 97.4 & 85.3 & 67.1 & 2.28 & 0.91 & 59.8 & 62.9 \\
    & - w/o execution    & 97.5 & 85.1 & 67.3 & 2.21 & 0.64 & 57.9 & 62.8 \\
    \hline
  \end{tabular}}

  \caption{\small
    Ablation study. We compare variants of \method-7B fine-tuned with different post-training stages and reward configurations.
  }
  \label{table:reward_ablation}
\end{table*}

%% file: tables/cold_start.tex
\begin{table*}[t!]
  \centering
  \begin{tabular}{lccccc}
    \hline
    \textbf{Model-Size} &
    \textbf{w/ SFT} &
    \textbf{w/ RL} &
    \textbf{Spider (Dev)} &
    \textbf{Spider (Test)} &
    \textbf{BIRD (Dev)} \\
    \hline
    \multirow{2}{*}{\method-3B} &
      \textcolor{green}{\checkmark} &
      \scalebox{0.8}{\textcolor{red}{\xmark}} &
      83.1 & 82.9 & 55.7 \\
    & \textcolor{green}{\checkmark} &
      \scalebox{0.8}{\textcolor{green}{\checkmark}} &
      84.6 & 84.3 & 55.2 \\
    \hline
    \multirow{3}{*}{\method-7B} &
      \textcolor{green}{\checkmark} &
      \textcolor{red}{\xmark} &
      83.6 & 83.5 & 58.7 \\
    & \scalebox{0.8}{\textcolor{red}{\xmark}} &
      \textcolor{green}{\checkmark} &
      85.8 & 86.8 & 61.7 \\
    & \textcolor{green}{\checkmark} &
      \textcolor{green}{\checkmark} &
      86.5 & 87.0 & 64.2 \\
    \hline
    \multirow{2}{*}{\method-14B} &
      \textcolor{green}{\checkmark} &
      \scalebox{0.8}{\textcolor{red}{\xmark}} &
      83.3 & 86.1 & 64.8 \\
    & \textcolor{green}{\checkmark} &
      \scalebox{0.8}{\textcolor{green}{\checkmark}} &
      87.1 & 88.5 & 66.7 \\
    \hline
  \end{tabular}
  \caption{\label{tab:cold_start}
    Execution accuracy (\%) of models with different cold-start strategies (SFT and RL).
  }
\end{table*}


%% file: custom.bib
@article{yu2018spider,
  title={Spider: A large-scale human-labeled dataset for complex and cross-domain semantic parsing and text-to-sql task},
  author={Yu, Tao and Zhang, Rui and Yang, Kai and Yasunaga, Michihiro and Wang, Dongxu and Li, Zifan and Ma, James and Li, Irene and Yao, Qingning and Roman, Shanelle and others},
  journal={arXiv preprint arXiv:1809.08887},
  year={2018}
}

@article{pourreza2024dts,
  title={Dts-sql: Decomposed text-to-sql with small large language models},
  author={Pourreza, Mohammadreza and Rafiei, Davood},
  journal={arXiv preprint arXiv:2402.01117},
  year={2024}
}

@inproceedings{
agarwal2024manyshot,
title={Many-shot In-Context Learning},
author={Rishabh Agarwal and Avi Singh and Lei M Zhang and Bernd Bohnet and Luis Rosias and Stephanie C.Y. Chan and Biao Zhang and Aleksandra Faust and Hugo Larochelle},
booktitle={ICML 2024 Workshop on In-Context Learning},
year={2024},
url={https://openreview.net/forum?id=goi7DFHlqS}
}

@article{wei2022chain,
  title={Chain-of-thought prompting elicits reasoning in large language models},
  author={Wei, Jason and Wang, Xuezhi and Schuurmans, Dale and Bosma, Maarten and Xia, Fei and Chi, Ed and Le, Quoc V and Zhou, Denny and others},
  journal={Advances in neural information processing systems},
  volume={35},
  pages={24824--24837},
  year={2022}
}

@article{hui2024qwen2,
      title={Qwen2. 5-Coder Technical Report},
      author={Hui, Binyuan and Yang, Jian and Cui, Zeyu and Yang, Jiaxi and Liu, Dayiheng and Zhang, Lei and Liu, Tianyu and Zhang, Jiajun and Yu, Bowen and Dang, Kai and others},
      journal={arXiv preprint arXiv:2409.12186},
      year={2024}
}

@article{li2024can,
  title={Can llm already serve as a database interface? a big bench for large-scale database grounded text-to-sqls},
  author={Li, Jinyang and Hui, Binyuan and Qu, Ge and Yang, Jiaxi and Li, Binhua and Li, Bowen and Wang, Bailin and Qin, Bowen and Geng, Ruiying and Huo, Nan and others},
  journal={Advances in Neural Information Processing Systems},
  volume={36},
  year={2024}
}

@article{snell2024scaling,
  title={Scaling llm test-time compute optimally can be more effective than scaling model parameters},
  author={Snell, Charlie and Lee, Jaehoon and Xu, Kelvin and Kumar, Aviral},
  journal={arXiv preprint arXiv:2408.03314},
  year={2024}
}

@article{pourreza2024din,
  title={Din-sql: Decomposed in-context learning of text-to-sql with self-correction},
  author={Pourreza, Mohammadreza and Rafiei, Davood},
  journal={Advances in Neural Information Processing Systems},
  volume={36},
  year={2024}
}

@article{wang2023mac,
  title={Mac-sql: Multi-agent collaboration for text-to-sql},
  author={Wang, Bing and Ren, Changyu and Yang, Jian and Liang, Xinnian and Bai, Jiaqi and Zhang, Qian-Wen and Yan, Zhao and Li, Zhoujun},
  journal={arXiv preprint arXiv:2312.11242},
  year={2023}
}

@article{sun2023sqlprompt,
  title={Sqlprompt: In-context text-to-sql with minimal labeled data},
  author={Sun, Ruoxi and Arik, Sercan {\"O} and Sinha, Rajarishi and Nakhost, Hootan and Dai, Hanjun and Yin, Pengcheng and Pfister, Tomas},
  journal={arXiv preprint arXiv:2311.02883},
  year={2023}
}

@inproceedings{li2023resdsql,
  title={Resdsql: Decoupling schema linking and skeleton parsing for text-to-sql},
  author={Li, Haoyang and Zhang, Jing and Li, Cuiping and Chen, Hong},
  booktitle={Proceedings of the AAAI Conference on Artificial Intelligence},
  volume={37},
  number={11},
  pages={13067--13075},
  year={2023}
}

@article{gan2021exploring,
  title={Exploring underexplored limitations of cross-domain text-to-SQL generalization},
  author={Gan, Yujian and Chen, Xinyun and Purver, Matthew},
  journal={arXiv preprint arXiv:2109.05157},
  year={2021}
}

@article{zhong2017seq2sql,
  title={Seq2sql: Generating structured queries from natural language using reinforcement learning},
  author={Zhong, Victor and Xiong, Caiming and Socher, Richard},
  journal={arXiv preprint arXiv:1709.00103},
  year={2017}
}

@article{qin2022survey,
  title={A Survey on Text-to-SQL Parsing: Concepts, Methods, and Future Directions},
  author={Qin, Bowen and Hui, Binyuan and Wang, Lihan and Yang, Min and Li, Jinyang and Li, Binhua and Geng, Ruiying and Cao, Rongyu and Sun, Jian and Si, Luo and others},
  journal={arXiv preprint arXiv:2208.13629},
  year={2022}
}

@article{xu2017sqlnet,
  title={Sqlnet: Generating structured queries from natural language without reinforcement learning},
  author={Xu, Xiaojun and Liu, Chang and Song, Dawn},
  journal={arXiv preprint arXiv:1711.04436},
  year={2017}
}

@article{scholak2021picard,
  title={PICARD: Parsing incrementally for constrained auto-regressive decoding from language models},
  author={Scholak, Torsten and Schucher, Nathan and Bahdanau, Dzmitry},
  journal={arXiv preprint arXiv:2109.05093},
  year={2021}
}

@article{c,
  title={Natural language interfaces to data},
  author={Quamar, Abdul and Efthymiou, Vasilis and Lei, Chuan and {\"O}zcan, Fatma and others},
  journal={Foundations and Trends{\textregistered} in Databases},
  volume={11},
  number={4},
  pages={319--414},
  year={2022},
  publisher={Now Publishers, Inc.}
}

@article{blier2021unbiased,
  title={Unbiased methods for multi-goal reinforcement learning},
  author={Blier, L{\'e}onard and Ollivier, Yann},
  journal={arXiv preprint arXiv:2106.08863},
  year={2021}
}

@article{shao2024deepseekmath,
  title={Deepseekmath: Pushing the limits of mathematical reasoning in open language models},
  author={Shao, Zhihong and Wang, Peiyi and Zhu, Qihao and Xu, Runxin and Song, Junxiao and Bi, Xiao and Zhang, Haowei and Zhang, Mingchuan and Li, YK and Wu, Y and others},
  journal={arXiv preprint arXiv:2402.03300},
  year={2024}
}

@article{wei2025swe,
  title={SWE-RL: Advancing LLM Reasoning via Reinforcement Learning on Open Software Evolution},
  author={Wei, Yuxiang and Duchenne, Olivier and Copet, Jade and Carbonneaux, Quentin and Zhang, Lingming and Fried, Daniel and Synnaeve, Gabriel and Singh, Rishabh and Wang, Sida I},
  journal={arXiv preprint arXiv:2502.18449},
  year={2025}
}

@article{jaech2024openai,
  title={Openai o1 system card},
  author={Jaech, Aaron and Kalai, Adam and Lerer, Adam and Richardson, Adam and El-Kishky, Ahmed and Low, Aiden and Helyar, Alec and Madry, Aleksander and Beutel, Alex and Carney, Alex and others},
  journal={arXiv preprint arXiv:2412.16720},
  year={2024}
}

@inproceedings{gan-etal-2021-towards,
    title = "Towards Robustness of Text-to-{SQL} Models against Synonym Substitution",
    author = "Gan, Yujian  and
      Chen, Xinyun  and
      Huang, Qiuping  and
      Purver, Matthew  and
      Woodward, John R.  and
      Xie, Jinxia  and
      Huang, Pengsheng",
    month = aug,
    year = "2021",
    address = "Online",
    publisher = "Association for Computational Linguistics",
    url = "https://aclanthology.org/2021.acl-long.195",
    doi = "10.18653/v1/2021.acl-long.195",
    pages = "2505--2515",
}

@misc{zhang2023actsqlincontextlearningtexttosql,
      title={ACT-SQL: In-Context Learning for Text-to-SQL with Automatically-Generated Chain-of-Thought}, 
      author={Hanchong Zhang and Ruisheng Cao and Lu Chen and Hongshen Xu and Kai Yu},
      year={2023},
      eprint={2310.17342},
      archivePrefix={arXiv},
      primaryClass={cs.CL},
      url={https://arxiv.org/abs/2310.17342}, 
}

@misc{liu2025skyrlsql,
      title={SkyRL-SQL: Matching GPT-4o and o4-mini on Text2SQL with Multi-Turn RL},
      author={Shu Liu and Sumanth Hegde and Shiyi Cao and Alan Zhu and Dacheng Li and Tyler Griggs and Eric Tang and Akshay Malik and Kourosh Hakhamaneshi and Richard Liaw and Philipp Moritz and Matei Zaharia and Joseph E. Gonzalez and Ion Stoica},
      year={2025},
      note={Notion Blog}
}

@misc{cao2025skyrl,
  title     = {SkyRL-v0: Train Real-World Long-Horizon Agents via Reinforcement Learning},
  author    = {Shiyi Cao and Sumanth Hegde and Dacheng Li and Tyler Griggs and Shu Liu and Eric Tang and Jiayi Pan and Xingyao Wang and Akshay Malik and Graham Neubig and Kourosh Hakhamaneshi and Richard Liaw and Philipp Moritz and Matei Zaharia and Joseph E. Gonzalez and Ion Stoica},
  year      = {2025},
}

@article{c3sql,
  title={C3: Zero-shot text-to-sql with chatgpt},
  author={Dong, Xuemei and Zhang, Chao and Ge, Yuhang and Mao, Yuren and Gao, Yunjun and Lin, Jinshu and Lou, Dongfang and others},
  journal={arXiv preprint arXiv:2307.07306},
  year={2023}
}

@article{openaigpt4,
title={GPT-4 Technical Report},
author={OpenAI},
journal={arXiv preprint arXiv:2303.08774},
year={2023}
}

@article{eyal2023semantic,
  title={Semantic Decomposition of Question and SQL for Text-to-SQL Parsing},
  author={Eyal, Ben and Bachar, Amir and Haroche, Ophir and Mahabi, Moran and Elhadad, Michael},
  journal={arXiv preprint arXiv:2310.13575},
  year={2023}
}

@article{petsql,
  title={PET-SQL: A Prompt-enhanced Two-stage Text-to-SQL Framework with Cross-consistency},
  author={Li, Zhishuai and Wang, Xiang and Zhao, Jingjing and Yang, Sun and Du, Guoqing and Hu, Xiaoru and Zhang, Bin and Ye, Yuxiao and Li, Ziyue and Zhao, Rui and others},
  journal={arXiv preprint arXiv:2403.09732},
  year={2024}
}

@article{dinsql,
title={DIN-SQL: Decomposed In-Context Learning of Text-to-SQL with Self-Correction},
author={Pourreza, Mohammadreza and Rafiei, Davood},
journal={arXiv preprint arXiv:2304.11015},
year={2023}
}

@misc{macsql,
title={MAC-SQL: A Multi-Agent Collaborative Framework for Text-to-SQL},
author={Bing Wang and Changyu Ren and Jian Yang and Xinnian Liang and Jiaqi Bai and Linzheng Chai and Zhao Yan and Qian-Wen Zhang and Di Yin and Xing Sun and Zhoujun Li},
year={2024},
eprint={2312.11242},
archivePrefix={arXiv},
primaryClass={cs.CL}
}

@article{omnisql,
  title={OmniSQL: Synthesizing High-quality Text-to-SQL Data at Scale},
  author={Li, Haoyang and Wu, Shang and Zhang, Xiaokang and Huang, Xinmei and Zhang, Jing and Jiang, Fuxin and Wang, Shuai and Zhang, Tieying and Chen, Jianjun and Shi, Rui and others},
  journal={arXiv preprint arXiv:2503.02240},
  year={2025}
}

@article{chasesql,
  title={Chase-sql: Multi-path reasoning and preference optimized candidate selection in text-to-sql},
  author={Pourreza, Mohammadreza and Li, Hailong and Sun, Ruoxi and Chung, Yeounoh and Talaei, Shayan and Kakkar, Gaurav Tarlok and Gan, Yu and Saberi, Amin and Ozcan, Fatma and Arik, Sercan O},
  journal={arXiv preprint arXiv:2410.01943},
  year={2024}
}

@article{deepseekr1,
  title={Deepseek-r1: Incentivizing reasoning capability in llms via reinforcement learning},
  author={Guo, Daya and Yang, Dejian and Zhang, Haowei and Song, Junxiao and Zhang, Ruoyu and Xu, Runxin and Zhu, Qihao and Ma, Shirong and Wang, Peiyi and Bi, Xiao and others},
  journal={arXiv preprint arXiv:2501.12948},
  year={2025}
}

@article{searchr1,
  title={Search-R1: Training LLMs to Reason and Leverage Search Engines with Reinforcement Learning},
  author={Jin, Bowen and Zeng, Hansi and Yue, Zhenrui and Wang, Dong and Zamani, Hamed and Han, Jiawei},
  journal={arXiv preprint arXiv:2503.09516},
  year={2025}
}

@article{grpo,
  title={Deepseekmath: Pushing the limits of mathematical reasoning in open language models},
  author={Shao, Zhihong and Wang, Peiyi and Zhu, Qihao and Xu, Runxin and Song, Junxiao and Bi, Xiao and Zhang, Haowei and Zhang, Mingchuan and Li, YK and Wu, Y and others},
  journal={arXiv preprint arXiv:2402.03300},
  year={2024}
}

@article{cao2024rsl,
  title={Rsl-sql: Robust schema linking in text-to-sql generation},
  author={Cao, Zhenbiao and Zheng, Yuanlei and Fan, Zhihao and Zhang, Xiaojin and Chen, Wei and Bai, Xiang},
  journal={arXiv preprint arXiv:2411.00073},
  year={2024}
}

@article{deepretrieval,
  title={DeepRetrieval: Hacking Real Search Engines and Retrievers with Large Language Models via Reinforcement Learning},
  author={Jiang, Pengcheng and Lin, Jiacheng and Cao, Lang and Tian, Runchu and Kang, SeongKu and Wang, Zifeng and Sun, Jimeng and Han, Jiawei},
  journal={arXiv preprint arXiv:2503.00223},
  year={2025}
}

@article{plaat2024reasoning,
  title={Reasoning with large language models, a survey},
  author={Plaat, Aske and Wong, Annie and Verberne, Suzan and Broekens, Joost and van Stein, Niki and Back, Thomas},
  journal={arXiv preprint arXiv:2407.11511},
  year={2024}
}

@article{qwen2.5,
  title={Qwen2. 5-coder technical report},
  author={Hui, Binyuan and Yang, Jian and Cui, Zeyu and Yang, Jiaxi and Liu, Dayiheng and Zhang, Lei and Liu, Tianyu and Zhang, Jiajun and Yu, Bowen and Lu, Keming and others},
  journal={arXiv preprint arXiv:2409.12186},
  year={2024}
}

@article{reasoningsql,
  title={Reasoning-SQL: Reinforcement Learning with SQL Tailored Partial Rewards for Reasoning-Enhanced Text-to-SQL},
  author={Pourreza, Mohammadreza and Talaei, Shayan and Sun, Ruoxi and Wan, Xingchen and Li, Hailong and Mirhoseini, Azalia and Saberi, Amin and Arik, Sercan and others},
  journal={arXiv preprint arXiv:2503.23157},
  year={2025}
}

@article{ma2025sqlr1trainingnaturallanguage,
      title={SQL-R1: Training Natural Language to SQL Reasoning Model By Reinforcement Learning}, 
      author={Peixian Ma and Xialie Zhuang and Chengjin Xu and Xuhui Jiang and Ran Chen and Jian Guo},
      year={2025},
      eprint={2504.08600},
      archivePrefix={arXiv},
      primaryClass={cs.DB},
      journal={arXiv preprint arXiv:2504.08600}, 
}

@article{liu2025survey,
  title={A Survey of Text-to-SQL in the Era of LLMs: Where are we, and where are we going?},
  author={Liu, Xinyu and Shen, Shuyu and Li, Boyan and Ma, Peixian and Jiang, Runzhi and Zhang, Yuxin and Fan, Ju and Li, Guoliang and Tang, Nan and Luo, Yuyu},
  journal={IEEE Transactions on Knowledge and Data Engineering},
  year={2025},
  publisher={IEEE}
}

@article{chen2025text,
  title={Text-to-SQL for Enterprise Data Analytics},
  author={Chen, Albert and Bundele, Manas and Ahlawat, Gaurav and Stetz, Patrick and Wang, Zhitao and Fei, Qiang and Jung, Donghoon and Chu, Audrey and Jayaraman, Bharadwaj and Panth, Ayushi and others},
  journal={arXiv preprint arXiv:2507.14372},
  year={2025}
}

@inproceedings{elgohary-etal-2020-speak,
    title = "Speak to your Parser: Interactive Text-to-{SQL} with Natural Language Feedback",
    author = "Elgohary, Ahmed  and
      Hosseini, Saghar  and
      Hassan Awadallah, Ahmed",
    editor = "Jurafsky, Dan  and
      Chai, Joyce  and
      Schluter, Natalie  and
      Tetreault, Joel",
    booktitle = "Proceedings of the 58th Annual Meeting of the Association for Computational Linguistics",
    month = jul,
    year = "2020",
    address = "Online",
    publisher = "Association for Computational Linguistics",
    url = "https://aclanthology.org/2020.acl-main.187/",
    doi = "10.18653/v1/2020.acl-main.187",
    pages = "2065--2077"
}

@misc{wei2025webagentr1trainingwebagents,
      title={WebAgent-R1: Training Web Agents via End-to-End Multi-Turn Reinforcement Learning}, 
      author={Zhepei Wei and Wenlin Yao and Yao Liu and Weizhi Zhang and Qin Lu and Liang Qiu and Changlong Yu and Puyang Xu and Chao Zhang and Bing Yin and Hyokun Yun and Lihong Li},
      year={2025},
      eprint={2505.16421},
      archivePrefix={arXiv},
      primaryClass={cs.CL},
      url={https://arxiv.org/abs/2505.16421}, 
}

@inproceedings{hu-etal-2025-os,
    title = "{OS} Agents: A Survey on {MLLM}-based Agents for Computer, Phone and Browser Use",
    author = "Hu, Xueyu  and
      Xiong, Tao  and
      Yi, Biao  and
      Wei, Zishu  and
      Xiao, Ruixuan  and
      Chen, Yurun  and
      Ye, Jiasheng  and
      Tao, Meiling  and
      Zhou, Xiangxin  and
      Zhao, Ziyu  and
      Li, Yuhuai  and
      Xu, Shengze  and
      Wang, Shenzhi  and
      Xu, Xinchen  and
      Qiao, Shuofei  and
      Wang, Zhaokai  and
      Kuang, Kun  and
      Zeng, Tieyong  and
      Wang, Liang  and
      Li, Jiwei  and
      Jiang, Yuchen Eleanor  and
      Zhou, Wangchunshu  and
      Wang, Guoyin  and
      Yin, Keting  and
      Zhao, Zhou  and
      Yang, Hongxia  and
      Wu, Fan  and
      Zhang, Shengyu  and
      Wu, Fei",
    editor = "Che, Wanxiang  and
      Nabende, Joyce  and
      Shutova, Ekaterina  and
      Pilehvar, Mohammad Taher",
    booktitle = "Proceedings of the 63rd Annual Meeting of the Association for Computational Linguistics (Volume 1: Long Papers)",
    month = jul,
    year = "2025",
    address = "Vienna, Austria",
    publisher = "Association for Computational Linguistics",
    url = "https://aclanthology.org/2025.acl-long.369/",
    doi = "10.18653/v1/2025.acl-long.369",
    pages = "7436--7465",
    ISBN = "979-8-89176-251-0"
}

@inproceedings{
yu2025dapo,
title={{DAPO}: An Open-Source {LLM} Reinforcement Learning System at Scale},
author={Qiying Yu and Zheng Zhang and Ruofei Zhu and Yufeng Yuan and Xiaochen Zuo and YuYue and Weinan Dai and Tiantian Fan and Gaohong Liu and Juncai Liu and LingJun Liu and Xin Liu and Haibin Lin and Zhiqi Lin and Bole Ma and Guangming Sheng and Yuxuan Tong and Chi Zhang and Mofan Zhang and Ru Zhang and Wang Zhang and Hang Zhu and Jinhua Zhu and Jiaze Chen and Jiangjie Chen and Chengyi Wang and Hongli Yu and Yuxuan Song and Xiangpeng Wei and Hao Zhou and Jingjing Liu and Wei-Ying Ma and Ya-Qin Zhang and Lin Yan and Yonghui Wu and Mingxuan Wang},
booktitle={The Thirty-ninth Annual Conference on Neural Information Processing Systems},
year={2025},
url={https://openreview.net/forum?id=2a36EMSSTp}
}

@article{li2025flow,
    title={In-the-Flow Agentic System Optimization for Effective Planning and Tool Use},
    author={Li, Zhuofeng and Zhang, Haoxiang and Han, Seungju and Liu, Sheng and Xie, Jianwen and Zhang, Yu and Choi, Yejin and Zou, James and Lu, Pan},
    journal={arXiv preprint arXiv:2510.05592},
    year={2025}
}

@misc{zheng2025deepresearcherscalingdeepresearch,
      title={DeepResearcher: Scaling Deep Research via Reinforcement Learning in Real-world Environments}, 
      author={Yuxiang Zheng and Dayuan Fu and Xiangkun Hu and Xiaojie Cai and Lyumanshan Ye and Pengrui Lu and Pengfei Liu},
      year={2025},
      eprint={2504.03160},
      archivePrefix={arXiv},
      primaryClass={cs.AI},
      url={https://arxiv.org/abs/2504.03160}, 
}

@inproceedings{yao2023react,
  title = {{ReAct}: Synergizing Reasoning and Acting in Language Models},
  author = {Yao, Shunyu and Zhao, Jeffrey and Yu, Dian and Du, Nan and Shafran, Izhak and Narasimhan, Karthik and Cao, Yuan},
  booktitle = {International Conference on Learning Representations (ICLR) },
  year = {2023},
  html = {https://arxiv.org/abs/2210.03629},
}

@misc{anthropic_claude_3.7_sonnet_system_card,
  author = {Anthropic},
  title = {{Claude 3.7 Sonnet System Card}},
  howpublished = {\url{https://www.anthropic.com/claude-3-7-sonnet-system-card}},
  year = {2025},
  note = {Accessed: November 24, 2025}
}

@inproceedings{deng-etal-2021-structure,
    title = "Structure-Grounded Pretraining for Text-to-{SQL}",
    author = "Deng, Xiang  and
      Awadallah, Ahmed Hassan  and
      Meek, Christopher  and
      Polozov, Oleksandr  and
      Sun, Huan  and
      Richardson, Matthew",
    editor = "Toutanova, Kristina  and
      Rumshisky, Anna  and
      Zettlemoyer, Luke  and
      Hakkani-Tur, Dilek  and
      Beltagy, Iz  and
      Bethard, Steven  and
      Cotterell, Ryan  and
      Chakraborty, Tanmoy  and
      Zhou, Yichao",
    booktitle = "Proceedings of the 2021 Conference of the North American Chapter of the Association for Computational Linguistics: Human Language Technologies",
    month = jun,
    year = "2021",
    address = "Online",
    publisher = "Association for Computational Linguistics",
    url = "https://aclanthology.org/2021.naacl-main.105/",
    doi = "10.18653/v1/2021.naacl-main.105",
    pages = "1337--1350",
}
